\documentclass[11pt]{article}

\usepackage[preprint]{acl}

\usepackage{times}
\usepackage{latexsym}

\usepackage[T1]{fontenc}

\usepackage[utf8]{inputenc}

\usepackage{microtype}

\usepackage{inconsolata}

\usepackage{graphicx}
\usepackage{amsmath}
\usepackage{booktabs}

\usepackage{algorithm}
\usepackage{algpseudocode}
\usepackage{xcolor}        %
\usepackage{amssymb}

\title{Don't Read Everything: A Curvature-Conditioned Query for Linear Attention}

\author{
  \textbf{Dong Le\textsuperscript{1}}
  \quad
  \textbf{Thong Nguyen\textsuperscript{2}}
  \quad
  \textbf{Cong-Duy Nguyen\textsuperscript{3}}
  \quad
  \textbf{Anh-Tuan Luu\textsuperscript{1,3}\thanks{Corresponding author.}}
\\
  \textsuperscript{1}Nanyang Technological University,
  \textsuperscript{2}National University of Singapore,
  \textsuperscript{3}VinUniversity
\\
  \texttt{leducdon001@e.ntu.edu.sg}
  \quad
  \texttt{thong.nguyen@u.nus.edu}
\\
  \texttt{duy.ntc@vinuni.edu.vn}
  \quad
  \texttt{anhtuan.luu@ntu.edu.sg}
}

\begin{document}
\maketitle
\begingroup
\renewcommand{\thefootnote}{\fnsymbol{footnote}}
\footnotetext[2]{\raggedright Code: \url{https://github.com/ledong0110/ccq}.}
\endgroup

\begin{abstract}
Linear attention reduces the quadratic cost of softmax attention by
maintaining a recurrent fast-weight state, but it consistently lags
on in-context retrieval and long-context tasks. Existing remedies
act on the write side of memory through gating, delta updates, or
kernel feature maps, but the read step is left unchanged: every past
key contributes additively to the output, so useful targets are
diluted by the bulk of stored vectors. We borrow one specific piece of
softmax's geometry to construct a cheap read-time contraction of
the query. A second-order Taylor expansion of the softmax
log-partition at the isotropic-attention point gives a local
quadratic model whose curvature coincides with the running key
covariance, a quantity that can be maintained with the same
recurrent/chunkwise mechanism as the linear-attention state. The
associated linear operator contracts the query along the
high-variance directions of memory before it reads the state. We call this mechanism
Curvature-Conditioned Query (CCQ). CCQ modifies only the read step
and is composable with any linear-attention backbone. Attached to
GLA and Gated DeltaNet, it improves perplexity, zero-shot
downstream accuracy, S-NIAH retrieval at and beyond the training
context, length-extrapolation perplexity from 4K to 20K, and
LongBench accuracy.
\end{abstract}

\section{Introduction}
The Transformer architecture~\citep{transformer} has become the default
choice for sequence modelling, but its softmax attention scales
quadratically with sequence length, which makes long-context training
and inference expensive. To mitigate this, linear
attention~\citep{linear_attn} replaces the softmax inner product with
a kernelized dot product, reframing the read as a linear RNN with a
matrix-valued state and reducing the cost to linear time and constant
cache.

Early linear-attention variants underperformed standard Transformers
on language modelling, but recent enhancements have closed much of
this gap. Data-dependent gating, used by GLA~\citep{gla},
RetNet~\citep{retnet}, RWKV~\citep{rwkv}, Mamba~\citep{mamba}, and
Mamba2~\citep{mamba2}, lets the state forget old entries adaptively.
The delta rule, revived by~\citet{schlag_deltanet} and parallelized
in~\citet{deltanet}, replaces the additive write with a soft
overwrite of the closest stored key, and Gated DeltaNet combines the
two~\citep{gated_deltanet}.
Despite this progress, linear-attention models still trail softmax on
in-context retrieval and long-context tasks~\citep{zoology,based,ruler},
where the ability to single out one stored item matters most.

We argue that this gap is not only about \emph{what} is written into
memory but also about \emph{how} it is read. Softmax attention couples
two operations in one expression: a key--query inner product, and a
log-sum-exp normalizer that allocates a fixed probability budget
across the matches. Linear attention preserves the inner product but
discards the normalizer. In the vanilla additive recurrence, past
contributions accumulate without explicit competition at read time,
so useful targets can be diluted by the bulk of stored vectors. The
remedies above act on the write side, deciding more carefully what
enters the recurrent state.
These methods cannot adapt the read to the distribution of keys
observed before the query arrives.

We propose a cheap read-time correction inspired by the local
curvature of the missing normalizer. Once the keys are fixed, the
log-sum-exp competition term depends only on the query, and a
second-order Taylor expansion at the isotropic-attention point shows
that its curvature is the running sample covariance of the keys.
Applying the associated linear contraction to the query yields a
cleaned query that has been pulled away from the dominant covariance
directions of memory before it multiplies the linear-attention
state. We call this mechanism Curvature-Conditioned Query (CCQ).

\noindent The contributions of this paper are as follows:
\begin{itemize}
    \item We identify the read step as a distinct, under-explored
    lever for closing the softmax--linear gap, complementary to the
    write-side remedies that dominate prior work.
    \item We propose CCQ, a curvature-aware linear contraction
    built from a single Taylor expansion of softmax's log-partition,
    that composes additively with any linear-attention write rule.
    \item Attached without modification to GLA and Gated DeltaNet,
    CCQ yields improvements in perplexity, zero-shot downstream
    accuracy, S-NIAH retrieval, and length-extrapolation perplexity
    at 500M and 1.3B parameters, and in LongBench accuracy at 1.3B.
\end{itemize}

\section{Preliminary}
\label{sec:preliminaries}
\subsection{Softmax Attention}
\label{sec:prelim_softmax}

Given a sequence of $d$-dimensional input vectors $x_1, \dots, x_T$, a
single-head Transformer~\citep{transformer} forms queries, keys, and
values by linear projections
\begin{equation}
q_t = W_Q x_t, \quad k_t = W_K x_t, \quad v_t = W_V x_t,
\end{equation}
with $W_Q, W_K, W_V \in \mathbb{R}^{d\times d}$, and reads from the
past via causal softmax attention:
\begin{equation}
a_{ti}
\;=\;
\frac{\exp(k_i^\top q_t)}{\sum_{j\le t}\exp(k_j^\top q_t)},
\qquad
o_t \;=\; \sum_{i\le t} a_{ti}\, v_i .
\label{eq:prelim_softmax}
\end{equation}
Equivalently,
\begin{equation}
\log a_{ti}
\;=\;
k_i^\top q_t
\;-\;
\log\sum_{j\le t}\exp(k_j^\top q_t).
\label{eq:prelim_softmax_log}
\end{equation}
This form separates the raw key--query score from the log-partition
term, exposing softmax as an inner-product score followed by a
data-dependent normalizer. We will use this view in
Section~\ref{sec:method} to motivate CCQ as a lightweight read-time
correction inspired by the local curvature of the second term.

\subsection{Linear Attention}
\label{sec:prelim_linattn}

Linear attention~\citep{linear_attn} replaces the exponential kernel
$\exp(k_i^\top q_t)$ in~\eqref{eq:prelim_softmax} with the dot product
of a feature map $\phi(k_i)^\top \phi(q_t)$, where
$\phi:\mathbb{R}^{d}\to\mathbb{R}^{n}$. Dropping the denominator and
taking $\phi$ to be the identity (the simplest and most common choice
in modern variants) factors the read into the recurrent form
\begin{equation}
S_t \;=\; S_{t-1} + v_t k_t^\top \;\in\; \mathbb{R}^{d_v\times d_k},
\qquad
o_t \;=\; S_t\, q_t .
\label{eq:prelim_linattn}
\end{equation}
The matrix $S_t$ acts as a fast-weight
memory~\citep{schmidhuber_fwp,schlag_deltanet}: every past key--value
pair is written into it once via an outer product and read once per
query through a single matrix-vector product. This swaps the
$\mathcal{O}(T^2)$ cost and $\mathcal{O}(T)$ KV cache of softmax for
$\mathcal{O}(T)$ training cost and an $\mathcal{O}(1)$ cache of fixed
size $d_v d_k$, at the price of a finite-rank approximation of the
full attention matrix.

The recurrence~\eqref{eq:prelim_linattn} is the additive baseline; the
\emph{backbone} is the choice of write update that replaces it. Gated
variants such as GLA~\citep{gla}, RetNet~\citep{retnet}, and
Mamba2~\citep{mamba2} insert a data-dependent decay before the outer
product; delta-rule variants such as
DeltaNet~\citep{schlag_deltanet,deltanet} and Gated
DeltaNet~\citep{gated_deltanet} replace the additive write with a soft
overwrite of the closest stored key. Throughout this paper we treat
the backbone as a black box: CCQ modifies only how the state is read
and is therefore compatible with any update of
the form $S_t = \mathrm{Update}(S_{t-1};\, k_t, v_t, \dots)$.

\section{Method}
\label{sec:method}
\subsection{Motivation: the missing competition of linear attention}
\label{sec:method_motivation}

Softmax attention~\citep{transformer} couples two operations in a
single closed-form expression: a key--query inner product
$k_j^\top q_t$ that scores each past key, and a log-partition
\begin{equation}
A_t(q_t) \;=\; \log\sum_{m\le t}\exp(k_m^\top q_t)
\label{eq:logZ}
\end{equation}
that normalizes those scores into a probability simplex. The second
piece imposes competition for a fixed total probability mass. For a
fixed query, $A_t$ shifts every log-score equally and does not change
their ranking; rather, it normalizes the aggregate mass, while its
curvature records how the stored keys vary across query directions.

Linear attention~\citep{linear_attn} drops the normalizer to gain
linear time and constant cache. Its recurrent fast-weight
form~\citep{schmidhuber_fwp,schlag_deltanet}
\begin{equation}
S_t \;=\; \sum_{j\le t} v_j k_j^\top,
\quad
o_t \;=\; S_t q_t
\label{eq:linear_attn}
\end{equation}
preserves the inner-product score but discards the competition: in
this vanilla recurrence, past contributions accumulate without
explicit competition at read time. Contributions along dominant
covariance directions therefore accumulate as a constant background
that the query cannot escape, and useful needles drown in the bulk of stored
vectors~\citep{zoology,based,ruler}. Rebuilding $A_t$ exactly would
cost the full attention matrix and erase the linear efficiency. The
question is whether the geometric information in its local curvature,
namely the variation of stored keys across query directions,
can be captured cheaply enough to fit inside a linear recurrence.

\subsection{From competition to a query-side correction}
\label{sec:method_ccq}

The key observation is that the competition term $A_t$ is a function
of $q_t$ alone once the keys are fixed. Its curvature provides a
geometric summary of how the stored keys vary across query directions.
Two ideas follow.

\paragraph{Idea 1: replace $A_t$ with its local quadratic shape.}
Expanding $A_t$ to second order around $q\!=\!0$ gives (see
Appendix~\ref{sec:app_taylor} for the derivation)
\begin{equation}
\begin{aligned}
A_t(q) \;=\;\; & \log t + \mu_t^\top q \\
               &+ \tfrac{1}{2}\, q^\top\!\bigl(\bar C_t - \mu_t\mu_t^\top\bigr) q
                 + \mathcal{O}(\|q\|^3),
\end{aligned}
\end{equation}
with $\mu_t = \tfrac{1}{t}\sum_{j\le t} k_j$, the running second
moment $\bar C_t = \tfrac{1}{t}\sum_{j\le t} k_j k_j^\top$, and the
running sample covariance
\begin{equation}
\Sigma_t \;\triangleq\; \bar C_t - \mu_t\mu_t^\top
            \;=\; \operatorname{Cov}_{\le t}[k].
\end{equation}
The constant $\log t$ and the linear term $\mu_t^\top q$ shift every
score uniformly and play no role in differentiating one key from
another. The quadratic Hessian $\Sigma_t$ has eigendirections aligned
with \emph{directions of high key spread}: top eigenvectors point
where stored keys vary the most around their mean.

\paragraph{Idea 2: contract the query along the local curvature, then read.}
A natural way to use this knowledge is to contract the query along
the directions in which $A_t$ rises before letting it interact with
the linear memory. The Hessian $\Sigma_t$ furnishes a curvature-aware
linear operator on the query space, and applying $I-\lambda_t\Sigma_t$
to $q_t$ gives the \emph{cleaned query}
\begin{equation}
\begin{aligned}
q_t^{\mathrm{clean}} &\;=\; (I - \lambda_t\,\Sigma_t)\, q_t
                      \;=\; q_t \;-\; \lambda_t\, \Sigma_t\, q_t, \\
\lambda_t &\;=\; \sigma\!\bigl(W_\lambda q_t + b_\lambda\bigr).
\end{aligned}
\label{eq:clean}
\end{equation}
Equivalently, for the quadratic curvature penalty
$R_t(q)=\tfrac{1}{2}q^\top\Sigma_tq$, this is one gradient step
$q_t-\lambda_t\nabla_qR_t(q_t)$. The scalar $\lambda_t \in (0,1)$ is
a query-conditioned gate that controls how aggressively to contract.
The Hessian determines the contraction geometry rather than a unique
finite-step operator. We use the direct gradient step because it
requires only one $\Sigma_t q_t$ product and preserves the simple
recurrent implementation.
The matrix $\Sigma_t$ is
maintainable by the same recurrent/chunkwise mechanism as the
backbone state through two running statistics: $\bar C_t$ has the form
of~\eqref{eq:linear_attn} with the value stream replaced by the key
stream and no decay, and $\mu_t$ is a running sum of keys. Together
they cost one additional $d_k\!\times\!d_k$ matrix and one
$d_k$-vector per head (see Appendix~\ref{sec:app_chunkwise}).

The linear-attention read is then performed unchanged, but with the
cleaned query:
\begin{equation}
o_t \;=\; S_t\, q_t^{\mathrm{clean}}
       \;=\; S_t q_t \;-\; \lambda_t\, S_t \Sigma_t\, q_t .
\label{eq:read}
\end{equation}
The first term is the original linear-attention output. The second is
a structured, context-conditioned correction that subtracts a read
along the high-variance directions of memory, weighted by $\lambda_t$.
Because $S_t$ is never modified, CCQ slots onto a backbone's read
without changing its write rule; we instantiate this with GLA and
Gated DeltaNet in Sec.~\ref{sec:exp_setup}.

\paragraph{What the correction actually penalizes.}
Substituting~\eqref{eq:clean} into the bilinear score reveals the
selection pattern in score space:
\begin{equation}
s_{tj}^{\mathrm{CCQ}}
\;=\; k_j^\top q_t \;-\; \lambda_t\, k_j^\top \Sigma_t\, q_t.
\label{eq:score_form}
\end{equation}
The first term is the original inner-product score; the second
penalizes a candidate key by the amount that $k_j$ and $q_t$ co-vary
in the empirical distribution of past keys. Equivalently,
$k_j^\top \Sigma_t q_t = \operatorname{Cov}_{\le t}[k_j^\top k,\, k^\top q_t]$,
so a key is penalized only when its alignment with a typical stored
key correlates with that stored key's alignment with the query. Keys
whose alignment with memory is uncorrelated with the query's read
direction are left untouched. CCQ therefore biases the read toward
directions that are less represented in the running key covariance,
favouring targets that lie off the dominant covariance directions.
The strength of this bias is set
per-token by the bounded gate $\lambda_t$, so the model can dial the
correction up where it helps and down where the standard inner-product
score is already sufficient.

\paragraph{When does CCQ help retrieval?}
On a toy retrieval model with $\Sigma_t \approx \rho u u^\top$
dominated by a single high-variance direction $u$, the CCQ-corrected
margin between a target key $k_\star$ (alignment
$\alpha = \langle k_\star, u\rangle$) and a distractor $k_d$
(alignment $\beta$) gains the term $\lambda\rho(\beta-\alpha)(u^\top
q)$ relative to the un-cleaned margin
(Eq.~\eqref{eq:margin_shift}, derived in
Appendix~\ref{sec:app_proposition}). CCQ therefore widens the
target margin whenever the target lies off the high-variance
direction \emph{and} the query overlaps with that direction, and
reduces to identity in the boundary cases (target collinear with
$u$, or query orthogonal to $u$). The diagnostic in
Appendix~\ref{sec:app_cov_validation} is consistent with the
favourable regime in the cases we probe: needle keys sit off the
high-variance subspace, putting the cleaning operator on the
configuration it was designed for.

\subsection{Properties of the construction}
\label{sec:method_properties}

We now make precise the local link between CCQ and softmax that
the rest of the paper relies on. Restricted to the
isotropic-attention point $q=0$, the Hessian of $A_t$ is the sample
covariance of the keys,
\begin{equation}
\nabla^2_q A_t(0) \;=\; \Sigma_t \;=\; \operatorname{Cov}_{\le t}[k],
\label{eq:hess}
\end{equation}
which is the classical exponential-family identity that the Hessian
of the log-partition equals the covariance of the sufficient
statistic~\citep{expfam_var}; the proof is reproduced in
Appendix~\ref{sec:app_taylor}. The cleaning operator
$I - \lambda_t \Sigma_t$ in~\eqref{eq:clean} is therefore a
curvature-aware contraction of the query along the local softmax
geometry: an eigendirection with covariance eigenvalue $\sigma_i$ is
shrunk by a factor $1-\lambda_t\sigma_i$, while directions orthogonal
to memory are passed through unchanged. Where
softmax uses the full nonlinear $A_t$ at every query, CCQ uses only
its second-order Taylor expansion at a single fixed point ($q=0$)
through running statistics that
fit into the same recurrent/chunkwise mechanism as the backbone state
(at the cost of one $d_k\!\times\!d_k$ matrix and one $d_k$-vector per
head).

The identity is exact at $q=0$; using the resulting marginal
covariance at trained queries is a heuristic geometric extension, not
a claim that the Taylor remainder is small. When attention
concentrates on a few keys, the exact query-weighted covariance can
differ from this global statistic. We test the assumption on trained
representations below and discuss this case further in the appendix.

This Hessian--covariance correspondence has three immediate
consequences. First, the construction is \emph{adaptive}: both the
suppression direction $\Sigma_t q_t$ and its strength $\lambda_t$ are
query-dependent, so CCQ produces a different cleaning geometry at
every token and every head while using only a pair of shared running
statistics per head. Second, it is \emph{local to the keys}: the
correction is built from second-order statistics of the keys, not
from the query distribution, so $\Sigma_t$ literally records how
memory is filling up irrespective of where the next query lands.
Third, it is \emph{decoupled from the write rule}: the cleaning acts
on $q_t$ only, so the state recurrence
$S_t = \mathrm{Update}(S_{t-1};\, k_t, v_t, \ldots)$ is untouched.
This makes the construction architecturally compatible with any
linear-attention write rule; we evaluate it on GLA and Gated DeltaNet.

\begin{figure}[t]
\centering
\includegraphics[width=\linewidth]{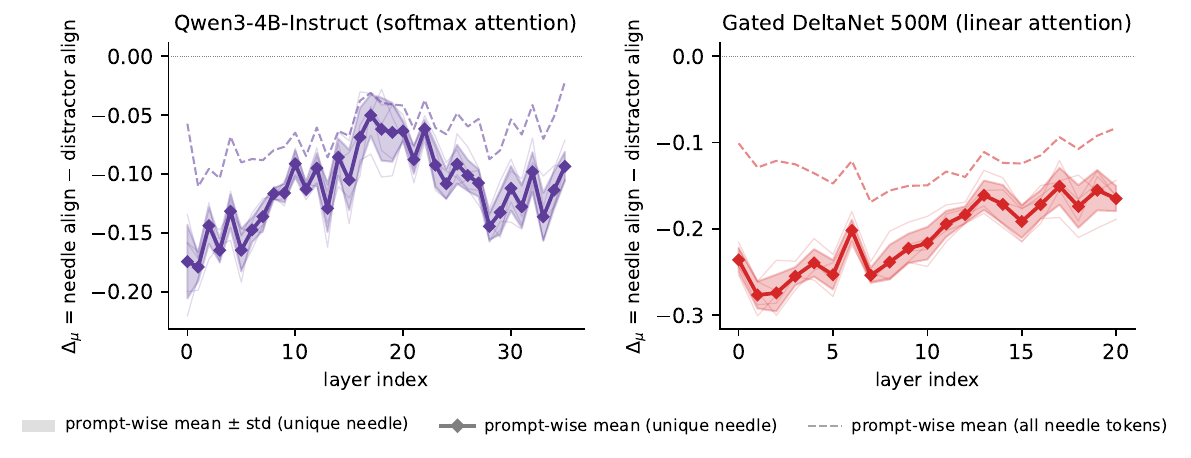}
\caption{Layer-by-layer geometric separation, aggregated over a
collection of needle-in-a-haystack prompts
(Appendix~\ref{sec:app_cov_validation}). Each curve plots
$\Delta_\mu = \text{needle align} - \text{distractor align}$ on the
top-16 eigenvector subspace of the centred key covariance $\Sigma$,
head-averaged. The solid line is the prompt-wise mean and the shaded
band is $\pm$ one prompt-wise standard deviation; the dashed line uses
all 17 needle-span tokens (including
common words like ``the'', ``is'') as a control. The unique-token
$\Delta$ is negative at \emph{every} layer of both models, with no
prompt-wise sign flip ($\sigma\!\approx\!0.02$). The effect is
strongest in early layers and weakens with depth, but never crosses
zero. The linear-attention model shows roughly $2\times$ the
separation of the softmax model at every depth; we report this as a
single-pair observation and do not claim a causal interpretation,
since the two models differ in scale, training data, and head
dimension as well as in attention type.}
\label{fig:cov_validation}
\end{figure}

\begin{figure}[t]
\centering
\includegraphics[width=\linewidth]{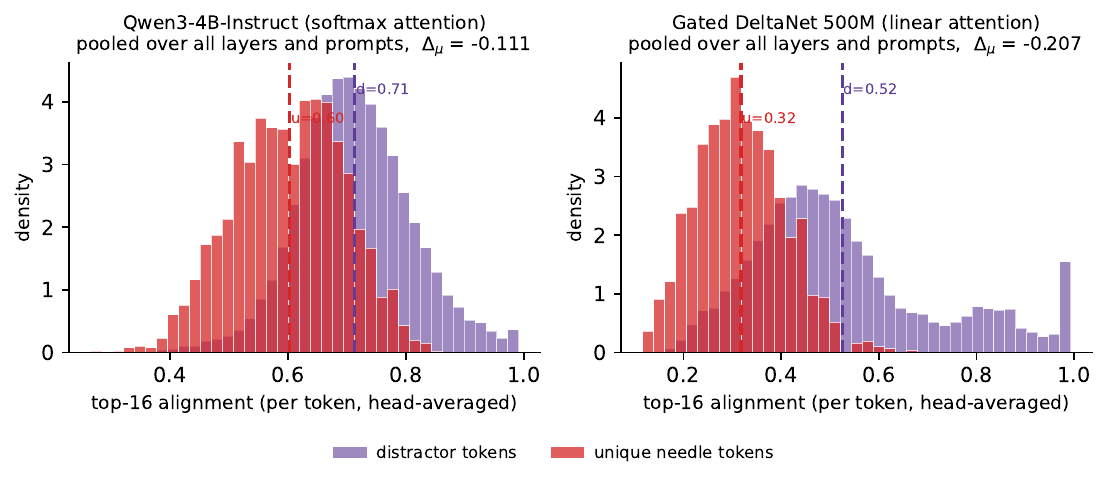}
\caption{Per-token distribution of top-16 alignment values pooled
over all layers and prompts. Distractor tokens (purple) form a
continuous distribution; the truly unique needle tokens (red
vertical bars) are shifted to the left of the distractor mean. The
pooled mean gaps are $-0.111$ (Qwen3) and $-0.207$ (Gated DeltaNet).}
\label{fig:cov_validation_distribution}
\end{figure}

\paragraph{Empirical check on the underlying assumption.}
The construction rests on the assumption that retrieval-relevant keys
live \emph{off} the high-variance directions of memory, while
distractor keys cluster inside them. We test this directly on
pretrained models by planting a unique needle sentence in a packed
context of $\sim$120 filler sentences and measuring, for the keys
captured at \emph{every} layer, the fraction of each key's centred
energy that lives in the top-16 eigenvector subspace of the running
key covariance $\Sigma$. To avoid single-prompt artefacts we average
across several distinct needle scenarios (different novel tokens,
sentences, and queries); see Appendix~\ref{sec:app_cov_validation} for
the full setup. On both Qwen3-4B-Instruct\footnote{\url{https://huggingface.co/Qwen/Qwen3-4B-Instruct-2507}} (softmax) and a
Gated DeltaNet 500M (linear attention), needle keys
project less onto the top-16 subspace than distractor keys at
\textbf{100\% of layers in both models}. The effect is consistent
across prompts (per-layer standard deviation $\sim$0.02) and stronger
in the linear-attention model
($\bar\Delta_\mu = -0.209\text{ vs }-0.111$, Fig.~\ref{fig:cov_validation}).
Pooling tokens across every layer and every prompt
(Fig.~\ref{fig:cov_validation_distribution}) gives a direct view of the
underlying distributions: the unique-needle histogram sits clearly to
the left of the distractor histogram in both models, so the per-layer
$\Delta_\mu$ in Fig.~\ref{fig:cov_validation} is not a thin
mean-of-means artefact but reflects a genuine population-level shift
of the needle keys away from the high-variance subspace.

\paragraph{Stability.}
With unit-norm keys and the sigmoid-gated $\lambda_t\!\in\!(0,1)$,
the cleaning operator $I - \lambda_t \Sigma_t$ has spectrum in
$[1-\lambda_t,1]\subset(0,1]$ at every position, independent of
sequence length, and acts
as a soft anisotropic contraction in the eigenbasis of $\Sigma_t$
that attenuates magnitude along high-variance directions without
annihilating any direction of $q_t$. The operator-norm bound and
spectrum argument are in Appendix~\ref{sec:app_stability}.

\section{Experiments}
\subsection{Experimental Setup}
\label{sec:exp_setup}

\paragraph{Training.}
We pretrain all models from scratch as autoregressive language models on
the same FineWeb-Edu~\citep{fineweb} subset, with the same tokenizer
(32K BPE), context length (4K), optimizer, and token budget within each
scale. We use AdamW with $\beta_1\!=\!0.9$, $\beta_2\!=\!0.95$, weight
decay $0.1$, peak learning rate $10^{-3}$, $1024$-step linear warmup,
and cosine decay to zero. Mixed precision is bfloat16. The global batch size is $524{,}288$
tokens per step at the 1.3B scale and $262{,}144$ tokens per step at
the 500M scale. We report two scales: at \emph{500M} every model is
trained for 15B tokens and at \emph{1.3B} for 40B tokens. The training
objective is standard next-token cross entropy with the fused norm
and fused cross-entropy kernels. Per-model layer counts, hidden
sizes, and token budgets are listed in Table~\ref{tab:model_configs};
the full optimizer settings are reproduced in
Table~\ref{tab:hparams} of Appendix~\ref{sec:app_hparams}.

\begin{table}[hbt]
\centering
\small
\setlength{\tabcolsep}{3pt}
\begin{tabular}{llrrr}
\toprule
Scale & Model & Layers & Hidden & Tokens \\
\midrule
500M & Transformer            & 35 & 1024 & 15B \\
500M & Mamba2                 & 65 & 1024 & 15B \\
500M & GLA                    & 21 & 1024 & 15B \\
500M & GLA-Hedgehog           & 21 & 1024 & 15B \\
500M & Gated DeltaNet         & 21 & 1024 & 15B \\
500M & CCQ-GLA                & 21 & 1024 & 15B \\
500M & CCQ-Gated DeltaNet     & 21 & 1024 & 15B \\
\midrule
1.3B & Transformer            & 24 & 2048 & 40B \\
1.3B & Mamba2                 & 48 & 2048 & 40B \\
1.3B & GLA                    & 21 & 2048 & 40B \\
1.3B & GLA-Hedgehog           & 21 & 2048 & 40B \\
1.3B & Gated DeltaNet         & 21 & 2048 & 40B \\
1.3B & CCQ-GLA                & 21 & 2048 & 40B \\
1.3B & CCQ-Gated DeltaNet     & 21 & 2048 & 40B \\
\bottomrule
\end{tabular}
\caption{Per-model layer counts, hidden sizes, and token budgets at
the 500M and 1.3B scales.}
\label{tab:model_configs}
\end{table}

\paragraph{Baselines.}
We compare CCQ to five backbones spanning softmax attention,
state-space, and linear-attention families: a Llama-style
\textbf{Transformer}, \textbf{Mamba2}~\citep{mamba2}, gated linear
attention (\textbf{GLA}~\citep{gla}), \textbf{GLA-Hedgehog} (GLA with
the Hedgehog learned-feature map~\citep{hedgehog}), and
\textbf{Gated DeltaNet}~\citep{gated_deltanet}. CCQ is attached on
top of GLA and Gated DeltaNet, giving \emph{CCQ-GLA} and
\emph{CCQ-Gated DeltaNet}; CCQ is applied at the read step. All
variants share SwiGLU MLPs, RMSNorm, untied input/output embeddings.

\paragraph{Evaluation.}
We probe four axes of model quality:
\emph{(i)} common-sense reasoning and language modelling
(\texttt{lm-evaluation-harness}~\citep{lm_eval_harness}, seven zero-shot
common-sense reasoning tasks plus LAMBADA and WikiText);
\emph{(ii)} synthetic in-context retrieval (three single-needle
tasks from the RULER evaluation suite~\citep{ruler} at 4K and 8K);
\emph{(iii)} length-extrapolation perplexity from 4K to 20K on six
long-context corpora; and
\emph{(iv)} long-context understanding (14-task English
LongBench~\citep{longbench}, zero-shot).
The per-family task selection, context lengths, metrics, and batch
sizes are listed in Appendix~\ref{sec:app_eval_details}.

\subsection{Language Modelling and Downstream Accuracy}
\label{sec:exp_lm}

Table~\ref{tab:results_lm} reports the language-modelling and
zero-shot downstream-accuracy numbers. At 500M, CCQ-Gated DeltaNet
attains the best WikiText perplexity (25.59) and the best 7-task
Avg (49.87\%) among the recurrent models, improving over its
backbone by $-0.74$ Wiki PPL and $+0.89$ Avg; CCQ-GLA improves over
GLA on LAMBADA accuracy ($+2.72$) and Wiki PPL ($-1.10$). At 1.3B,
CCQ-GLA attains the best WikiText perplexity (16.92), LAMBADA
perplexity (11.39), LAMBADA accuracy (47.82), and HellaSwag (59.14),
improving on its GLA backbone by $-1.88$ Wiki PPL, $-0.92$ LMB PPL,
and $+3.70$ HellaSwag. CCQ-Gated DeltaNet attains the best 7-task
Avg (56.24) and best SIQA (41.71), improving on its Gated DeltaNet
backbone by $+0.60$ Avg, with CCQ-GLA a close second on Avg
(56.10).

\begin{table*}[t]
\centering
\scriptsize
\setlength{\tabcolsep}{3pt}
\begin{tabular*}{\linewidth}{@{\extracolsep{\fill}}llrrrrrrrrrrr}
\toprule
Scale & Model & Wiki & LMB & LMB & PIQA & Hella. & Wino. & ARC-e & ARC-c & SIQA & BoolQ & Avg. \\
& & PPL$\downarrow$ & PPL$\downarrow$ & acc$\uparrow$ & acc & acc\_n & acc & acc\_n & acc\_n & acc & acc & acc \\
\midrule
\multicolumn{13}{l}{\emph{500M parameters / 15B tokens}} \\
\multicolumn{13}{l}{\emph{Recurrent models}} \\
500M & Mamba2                 &  26.22 & 29.37 & 34.62 & 67.30 & 44.15 & 52.57 & 55.13 & 29.27 & 39.92 & 55.84 & 49.17 \\
500M & GLA                    &  27.27 & 32.11 & 34.02 & 66.38 & 41.93 & 52.57 & 52.23 & 28.58 & 38.43 & 59.66 & 48.54 \\
500M & GLA-Hedgehog           &  32.47 & 43.74 & 29.50 & 66.21 & 40.48 & 51.85 & 53.37 & 28.33 & 39.41 & \textbf{61.71} & 48.77 \\
500M & Gated DeltaNet         &  26.33 & 29.10 & 33.88 & 67.85 & 43.48 & \textbf{53.67} & 55.56 & 28.67 & 39.41 & 54.19 & 48.98 \\
500M & \textbf{CCQ-GLA}       &  26.17 & 25.77 & 36.74 & 68.01 & 43.45 & 53.35 & \textbf{56.82} & 29.78 & 39.87 & 56.85 & 49.73 \\
500M & \textbf{CCQ-Gated DeltaNet} & \textbf{25.59} & \textbf{22.93} & \textbf{38.19} & \textbf{68.28} & \textbf{44.61} & 53.35 & 56.14 & \textbf{30.55} & \textbf{40.23} & 55.96 & \textbf{49.87} \\
\multicolumn{13}{l}{\emph{Attention models}} \\
500M & Transformer            & 331.69 & 30.47 & 34.04 & 67.03 & 43.12 & 54.93 & 54.80 & 30.12 & 38.08 & 58.01 & 49.44 \\
\midrule
\multicolumn{13}{l}{\emph{1.3B parameters / 40B tokens}} \\
\multicolumn{13}{l}{\emph{Recurrent models}} \\
1.3B & Mamba2                      & 18.17 & 13.55 & 45.55 & 72.58 & 57.00 & 56.91 & 64.60 & 36.52 & 41.61 & 56.97 & 55.17 \\
1.3B & GLA                         & 18.80 & 12.31 & 47.02 & 72.52 & 55.44 & \textbf{59.67} & 66.54 & \textbf{38.91} & 39.15 & 58.01 & 55.75 \\
1.3B & GLA-Hedgehog                & 19.07 & 15.17 & 43.08 & \textbf{72.91} & 55.90 & 56.43 & \textbf{67.76} & 38.65 & 40.69 & 59.17 & 55.93 \\
1.3B & Gated DeltaNet              & 17.98 & 13.11 & 45.57 & 71.27 & 56.66 & 57.22 & 65.91 & 36.35 & 41.45 & \textbf{60.58} & 55.64 \\
1.3B & \textbf{CCQ-GLA}            & \textbf{16.92} & \textbf{11.39} & \textbf{47.82} & 72.03 & \textbf{59.14} & 59.43 & 64.23 & 36.95 & 41.20 & 59.69 & 56.10 \\
1.3B & \textbf{CCQ-Gated DeltaNet} & 18.01 & 12.52 & 46.13 & 71.33 & 56.82 & 58.17 & 66.46 & 38.82 & \textbf{41.71} & 60.37 & \textbf{56.24} \\
\multicolumn{13}{l}{\emph{Attention models}} \\
1.3B & Transformer                 & 41.55 & 14.31 & 45.02 & 71.22 & 53.44 & 56.27 & 61.49 & 36.01 & 42.58 & 61.16 & 54.60 \\
\bottomrule
\end{tabular*}
\caption{Language modelling perplexity and zero-shot common-sense
reasoning accuracy. Hella.: HellaSwag; Wino.: WinoGrande.
\emph{acc\_n} denotes normalized accuracy where applicable. The Avg
column is the unweighted mean over PIQA, Hella., Wino., ARC-e, ARC-c,
SIQA, BoolQ (seven tasks); rows missing any task are reported as
\texttt{---}. \textbf{Bold} marks the best recurrent number per scale.
Cells marked --- were not evaluated.}
\label{tab:results_lm}
\end{table*}

\subsection{Synthetic Needle-in-a-Haystack}
\label{sec:exp_s_niah}

Table~\ref{tab:s_niah} resolves the single-needle tasks by context
length: \emph{S-NIAH-1} (pass-key), \emph{S-NIAH-2} (number), and
\emph{S-NIAH-3} (uuid). 1K/2K is in-distribution, 8K is pure length
extrapolation; S-NIAH-3 collapses past 4K for every model.
Difficulty grows from S-NIAH-1 to S-NIAH-3 as the needle overlaps
more with distractor tokens, and within each task as context
grows.

CCQ rescues both backbones at the lengths where they fail. At
500M, GLA collapses on S-NIAH-1 past 1K but CCQ-GLA recovers to
$97.6/56.4/20.6$ at 2K/4K/8K, and CCQ-Gated DeltaNet takes the best
8K columns on S-NIAH-2 ($+26$ over backbone) and S-NIAH-3 4K. At
1.3B the gains sharpen: CCQ-GLA adds $+49.0$ on S-NIAH-1 4K,
$+29.4$ on S-NIAH-2 8K, and $+35.6$ on S-NIAH-3 4K; CCQ-Gated
DeltaNet adds $+40.6$ on S-NIAH-3 4K, and CCQ-GLA wins S-NIAH-2
4K/8K and S-NIAH-3 1K across \emph{all seven models}. The
kernel-feature-map baseline GLA-Hedgehog improves GLA on S-NIAH-1 at
2K/4K but degrades most other cells, showing that a generic
inner-product change does not reproduce CCQ's pattern of gains. The
Transformer leads in-distribution S-NIAH-3 at 1.3B but collapses to
0 at 8K, where the recurrent CCQ variants retain a substantial
fraction of in-distribution accuracy.

\begin{table*}[hbt]
\centering
\scriptsize
\setlength{\tabcolsep}{4pt}
\begin{tabular*}{\linewidth}{@{\extracolsep{\fill}}ll cccc cccc ccc}
\toprule
& & \multicolumn{4}{c}{S-NIAH-1 (pass-key)}
  & \multicolumn{4}{c}{S-NIAH-2 (number)}
  & \multicolumn{3}{c}{S-NIAH-3 (uuid)} \\
\cmidrule(lr){3-6}\cmidrule(lr){7-10}\cmidrule(lr){11-13}
Scale & Model & 1K & 2K & 4K & 8K & 1K & 2K & 4K & 8K & 1K & 2K & 4K \\
\midrule
500M & Mamba2                       & \textbf{100.0} & \textbf{100.0} & 97.6           & 68.4           & \textbf{100.0} & \textbf{100.0} & \textbf{97.6} & 45.2           & \textbf{68.6} & \textbf{54.8} & 28.8 \\
500M & GLA                          & 88.0           & 26.2           & 11.0           &  5.2           & 99.8           & 94.8           & 41.6           & 13.8           & 39.8 & 16.0 &  4.2 \\
500M & GLA-Hedgehog                 & 76.6           & 55.8           & 36.6           &  0.0           & 93.8           & 85.6           & 16.4           &  0.0           & 11.8 &  3.4 &  0.4 \\
500M & Gated DeltaNet               & \textbf{100.0} & \textbf{100.0} & \textbf{100.0} & \textbf{100.0} & \textbf{100.0} & \textbf{100.0} & 84.6           & 24.8           & 56.2 & 54.2 & 30.2 \\
500M & \textbf{CCQ-GLA}             & \textbf{100.0} & 97.6           & 56.4           & 20.6           & 99.8           & \textbf{100.0} & 93.6           & 27.8           & 11.0 & 41.4 &  3.6 \\
500M & \textbf{CCQ-Gated DeltaNet}  & \textbf{100.0} & \textbf{100.0} & \textbf{100.0}           & 99.6           & \textbf{100.0} & 99.0           & 87.0           & \textbf{50.8}  & 67.6 & 53.4 & \textbf{33.6} \\
500M & Transformer                 & \textbf{100.0} & \textbf{100.0} & 97.8           &  0.0           & 40.2           & 98.2           & 91.2           &  0.0           & 12.2 &  7.2 & 12.6 \\
\midrule
1.3B & Mamba2                      &           99.8 & \textbf{100.0} & \textbf{100.0} & 99.2           & \textbf{100.0} & \textbf{100.0} & 79.6           & 55.8           & 69.4           & 47.6           & 27.4           \\
1.3B & GLA                         & \textbf{100.0} & 86.4           & 47.8           & 13.0           & \textbf{100.0} & 99.4           & 93.6           & 31.2           & 66.0           & 51.6           & 29.2           \\
1.3B & GLA-Hedgehog                & 92.4           & 96.0           & 75.0           &  6.2           & 99.8           & 98.4           & 62.4           &  3.6           & 49.2           & 23.8           &  1.8           \\
1.3B & Gated DeltaNet              & \textbf{100.0} & 99.8           & \textbf{100.0} & \textbf{99.8}  & \textbf{100.0} & \textbf{100.0} & 83.4           & 56.6  & 64.4           & 56.0           &  6.0           \\
1.3B & \textbf{CCQ-GLA}            & \textbf{100.0} & \textbf{100.0} & 96.8           & 42.6           & \textbf{100.0} & \textbf{100.0} & \textbf{99.4} & \textbf{60.6}  & \textbf{94.4} & 90.4           & 64.8           \\
1.3B & \textbf{CCQ-Gated DeltaNet} & \textbf{100.0} & 99.8           & \textbf{100.0} & 98.2           & \textbf{100.0} & \textbf{100.0} & 86.0           & 58.4           & 66.4           & 59.2           & 46.6           \\
1.3B & Transformer                 & \textbf{100.0} & \textbf{100.0} & 99.8           &  0.0           & 99.8           & \textbf{100.0} & 99.0           &  0.0           & 93.0           & \textbf{94.4}  & \textbf{82.2}  \\
\bottomrule
\end{tabular*}
\caption{Synthetic needle-in-a-haystack accuracy as a function of
context length, on the 500M and 1.3B checkpoints. S-NIAH-1: pass-key
retrieval. S-NIAH-2: number in haystack. S-NIAH-3: uuid in haystack
(reported up to 4K; uuid retrieval collapses past that for every
model we tested). All values are percentages. \textbf{Bold} marks
the best per column at each scale. The 8K columns probe pure length
extrapolation past the 4K training window; the Transformer's
collapse at 8K is expected since its training context is only 4K.}
\label{tab:s_niah}
\end{table*}

\subsection{Length Extrapolation}
\label{sec:exp_length_extrap}

We compute token-level perplexity on a 4K-to-20K sweep in 2K steps
(nine points) over six long-context corpora (GovReport, QMSum,
NarrativeQA, Qasper, PG19, CodeParrot); since all models are trained
at 4K, the 6K--20K points are pure extrapolation
(Appendix~\ref{sec:app_eval_details}).

\begin{figure}[t]
\centering
\includegraphics[width=\linewidth]{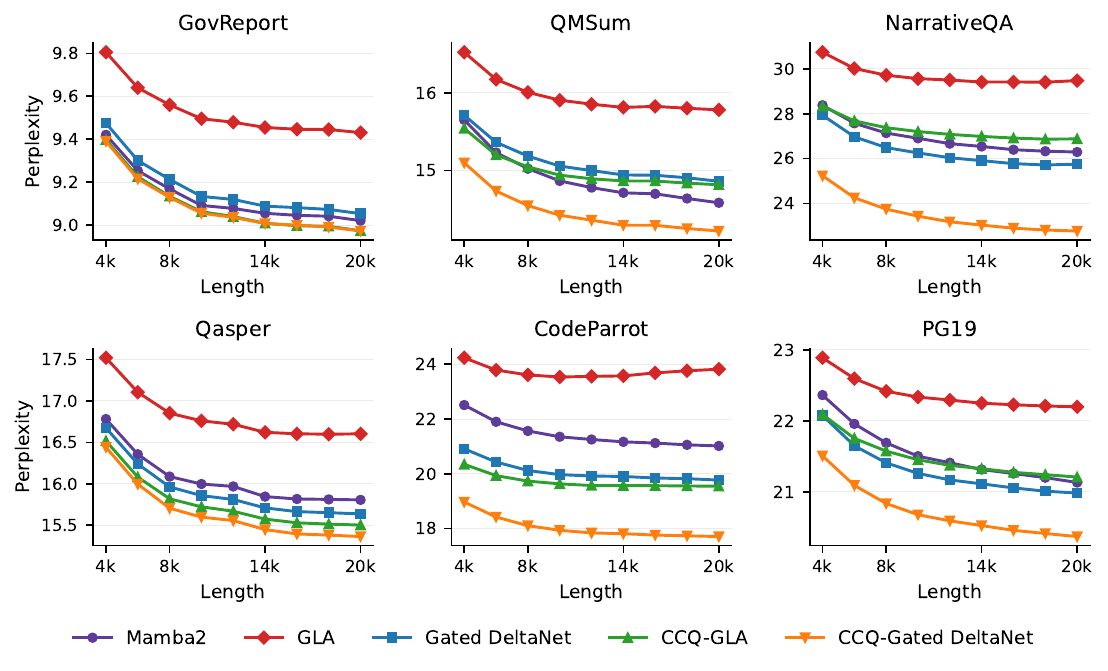}
\caption{Length-extrapolation PPL at 500M over six corpora and nine
lengths (4K--20K in 2K steps). Lower is better. Five recurrent
models shown (Mamba2, GLA, Gated DeltaNet, CCQ-GLA, CCQ-Gated
DeltaNet); the Transformer and GLA-Hedgehog are off-range and
omitted. Y-axes are per-panel auto-scaled to the shown curves.}
\label{fig:length_extrap_500M}
\end{figure}

At 500M (Fig.~\ref{fig:length_extrap_500M}), CCQ-Gated DeltaNet
attains the lowest PPL on every dataset at every length, with the
margin over its backbone widening as context grows; CCQ-GLA sits
well below GLA, closing most of the gap to the stronger recurrent
baselines. The Transformer (4K-only training) and GLA-Hedgehog
(steeply rising curves) are off-range and not plotted.

\begin{figure}[t]
\centering
\includegraphics[width=\linewidth]{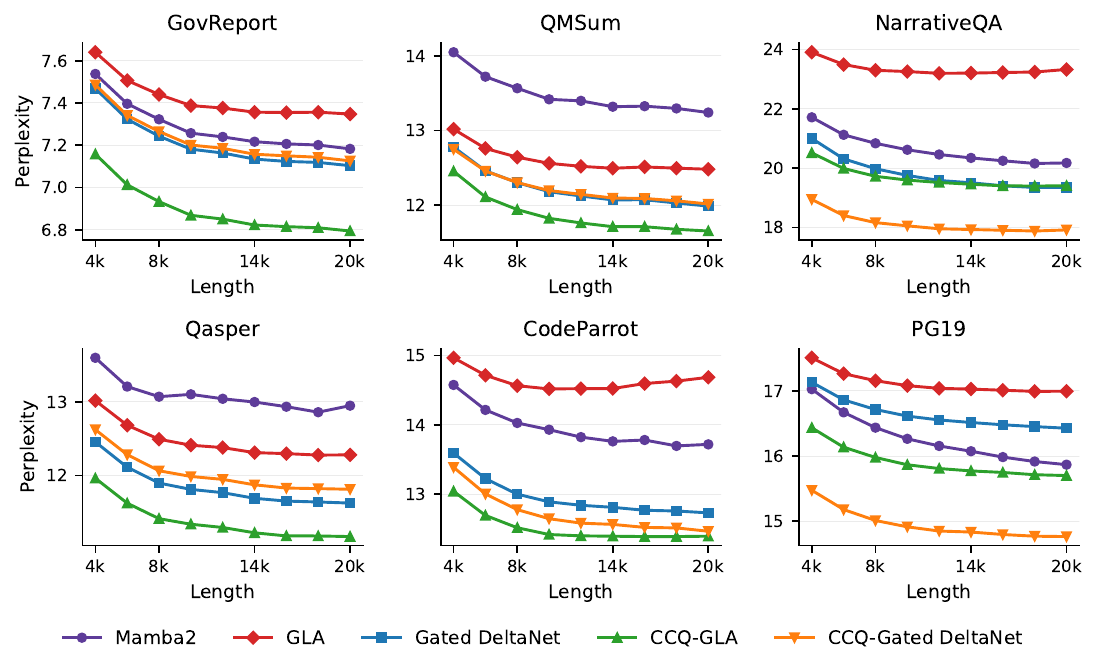}
\caption{Length-extrapolation PPL at 1.3B on the same six corpora
and nine lengths as Fig.~\ref{fig:length_extrap_500M}. Five
recurrent models shown; Transformer and GLA-Hedgehog are off-range
and omitted.}
\label{fig:length_extrap_1B}
\end{figure}

At 1.3B (Fig.~\ref{fig:length_extrap_1B}) the two CCQ variants
split the panels at 20K: CCQ-GLA wins GovReport, QMSum, Qasper,
and CodeParrot; CCQ-Gated DeltaNet wins NarrativeQA and PG19. Each
CCQ variant beats its own backbone on every panel by $1$--$2.5$
PPL at 20K. The Transformer 1.3B is off-range (CodeParrot 20K
$>\!1700$).

\subsection{Long-Context Understanding}
\label{sec:exp_longbench}

Table~\ref{tab:longbench} reports zero-shot accuracy on the 14
English LongBench tasks. We only report at the 1.3B scale
(Appendix~\ref{sec:app_eval_details}). CCQ-GLA attains the best
14-task Avg (15.77) among the recurrent models, improving over its
GLA backbone by $+1.70$ Avg, with CCQ-Gated DeltaNet a close second
(14.78, $+0.98$ over its own backbone) and ahead of Mamba2 by
$+2.64$. CCQ-GLA leads the QA columns (Qasper, MFQ, 2WM) and the
GovReport/TREC/RepoBench-P axes; CCQ-Gated DeltaNet leads
MultiNews, TriviaQA, and SAMSum. Both CCQ variants improve on their
respective backbones across the longer-output tasks (GovReport,
MultiNews, SAMSum) and on the few-shot TREC/TriviaQA columns by
$4$--$14$ points. The Transformer baseline collapses on every
long-input column (Avg $4.80$) because its $4{,}096$-token training
window is much shorter than the LongBench inputs.

\begin{table*}[hbt]
\centering
\scriptsize
\setlength{\tabcolsep}{4pt}
\begin{tabular*}{\linewidth}{@{\extracolsep{\fill}}ll ccc ccc ccc ccc cc r}
\toprule
& & \multicolumn{3}{c}{Single-Doc QA}
  & \multicolumn{3}{c}{Multi-Doc QA}
  & \multicolumn{3}{c}{Summarization}
  & \multicolumn{3}{c}{Few-shot}
  & \multicolumn{2}{c}{Code}
  & \\
\cmidrule(lr){3-5}\cmidrule(lr){6-8}\cmidrule(lr){9-11}\cmidrule(lr){12-14}\cmidrule(lr){15-16}
Scale & Model
 & NQA & QQA & MFQ & HQA & 2WM & Mus
 & GvR & QMS & MNs & TRC & TQA & SSM & LCC & RBP & Avg. \\
\midrule
\multicolumn{17}{l}{\emph{1.3B parameters / 40B tokens}} \\
\multicolumn{17}{l}{\emph{Recurrent models}} \\
1.3B & Mamba2                      & \textbf{2.29} & 5.75 & 13.30 & \textbf{6.03} &  7.51 &  2.90 &  7.39 & \textbf{18.13} & 11.10 & 36.00 & 22.25 &  8.52 & 19.25 & 23.33 & 13.13 \\
1.3B & GLA                         &  2.20 &  4.78 & 13.51 &  4.94 & 9.32 & \textbf{3.32} &  7.46 & 16.63 & 11.20 & 24.75 & 22.85 & 23.88 & 25.26 & 26.81 & 14.07 \\
1.3B & GLA-Hedgehog                &  1.08 &  3.47 &  9.95 &  3.91 &  4.70 &  2.22 &  7.03 & 11.06 &  9.03 &  8.00 & 19.31 &  7.88 & 18.52 & 18.96 &  8.94 \\
1.3B & Gated DeltaNet              &  2.12 &  5.28 & 13.43 &  5.77 &  7.78 &  3.12 &  7.11 & 17.39 & 11.33 & 23.00 & 21.01 & 25.68 & \textbf{24.62} & 25.60 & 13.80 \\
1.3B & \textbf{CCQ-GLA}            &  2.08 & \textbf{6.10} & \textbf{14.81} &  5.18 & \textbf{9.90} & 3.27 & \textbf{9.11} & 18.10 & 12.28 & \textbf{39.50} & 26.09 & 24.79 & 23.26 & \textbf{26.37} & \textbf{15.77} \\
1.3B & \textbf{CCQ-Gated DeltaNet} &  2.24 &  3.91 & 13.26 &  3.39 &  4.63 &  2.51 & 8.76 & 17.99 & \textbf{12.76} & 36.50 & \textbf{28.28} & \textbf{27.75} & 23.68 & 21.29 & 14.78 \\
\multicolumn{17}{l}{\emph{Attention models}} \\
1.3B & Transformer                 &  0.28 &  2.58 &  6.61 &  0.66 &  1.55 &  0.00 &  3.47 &  3.65 &  7.54 &  6.50 &  3.92 &  2.56 & 14.41 & 13.41 &  4.80 \\
\bottomrule
\end{tabular*}
\caption{LongBench accuracy on 14 English tasks, grouped by task
family, at the 1.3B scale. The 500M models are omitted because at
that scale the per-task standard error is comparable to the spread
between models and the ranking is dominated by noise. The Avg column
is the unweighted mean over the 14 tasks.
NQA: NarrativeQA. QQA: Qasper. MFQ: MultiField QA. HQA: HotpotQA.
2WM: 2WikiMultiQA. Mus: MuSiQue. GvR: GovReport. QMS: QMSum.
MNs: MultiNews. TRC: TREC. TQA: TriviaQA. SSM: SAMSum.
LCC: LCC. RBP: RepoBench-P.
All values are percentages; \textbf{bold} marks the best recurrent
number.}
\label{tab:longbench}
\end{table*}

\section{Related Works}
\label{sec:related_work}
Linear attention~\citep{linear_attn} replaces the $\mathcal{O}(T^2)$
pairwise softmax of the Transformer~\citep{transformer} with a
recurrent fast-weight memory~\citep{schmidhuber_fwp,schlag_deltanet},
yielding $\mathcal{O}(T)$ training and $\mathcal{O}(1)$ cache
inference. The Hopfield perspective~\citep{ramsauer_hopfield} makes
the limitation visible: vanilla linear attention writes with a
Hebbian rule whose capacity is bounded, which shows up as a recall
gap against softmax~\citep{zoology,based,ruler}. Existing remedies
group by which part of the read or write pipeline they modify.

\paragraph{Data-dependent gating.}
GLA~\citep{gla} adds a learned per-key forget gate; RetNet~\citep{retnet}
uses fixed exponential decay; RWKV~\citep{rwkv} interleaves a time
mixing decay with a channel-mixing MLP; Mamba~\citep{mamba} and
Mamba2~\citep{mamba2} cast the decay as a selective state-space
model conditioned on the input; xLSTM~\citep{xlstm} revisits LSTM
gating with matrix-valued cells. These methods control \emph{what
stays} in the state, but the read remains a single matrix--vector
product.

\paragraph{Delta rule and its gated extension.}
A second line replaces the additive write with a soft overwrite of
the matched key. The delta rule was revived
by~\citet{schlag_deltanet} and parallelised across sequence length
by~\citet{deltanet}. Gated DeltaNet~\citep{gated_deltanet} combines
a forget gate with the delta update and closes most of the
recall--retention gap. The delta family has stronger capacity than
Hebbian writes but still reads through the unmodified inner product.

\paragraph{Online-learning view of the write.}
A recent line interprets the recurrent state as the solution to a
small online learning problem and refines the write accordingly:
MesaNet~\citep{mesanet} solves an online least-squares problem at
each step, TTT~\citep{ttt} parameterises the state by a small
nonlinear predictor updated at test time, and Titans~\citep{titans}
adds a long-term memory module trained at inference. These deliver
sharper writes at the cost of nonlinear updates or per-token
inner-loop work, and again leave the read step unchanged.

\paragraph{Kernel feature maps.}
A third group changes the similarity function before any write:
Performer~\citep{performer} uses random features to approximate the
softmax kernel; RFA~\citep{rfa} applies a similar idea with
explicit normalisation; T2R~\citep{t2r} converts a pretrained
Transformer into a linear RNN by learning a feature map;
Hedgehog~\citep{hedgehog} learns an MLP feature map that mimics the
softmax similarity to recover the spiky kernel linear attention
typically loses. These reshape how keys and queries are compared
but still use a single inner-product read against the state.

\paragraph{Softmax expansions and higher-order states.}
\citet{mongaras_softmax} derive a recurrent form of softmax attention
from its Taylor expansion and show that linear attention corresponds
to its first-order term. Building on this view, 2Mamba2Furious
introduces a second-order hidden state and a modified $A$-mask for
Mamba-2~\citep{mamba2furious}. CCQ instead leaves the backbone state
update unchanged and uses second-order key statistics to modify the
query at read time.

\paragraph{Position of CCQ.}
CCQ acts only at the read step. The write
recurrence is left exactly as the chosen backbone defines it, and
only the query used to read from memory is modified, using both
the current query and the running second-order statistics of the
stored keys. The cleaning is therefore memory-aware in a way that
input-only query projections cannot be. Because it touches
only the read, CCQ composes additively with any of the write-side
mechanisms above; in our experiments we attach it without
modification to GLA and Gated
DeltaNet.

\section{Conclusion}
We introduced CCQ, a read-side correction that contracts the query
along the dominant covariance directions of memory using the running key
covariance, read off the local Hessian of softmax's log-partition
at the isotropic-attention point. Attached without modification
to GLA and Gated DeltaNet, CCQ achieves gains in perplexity,
downstream accuracy, S-NIAH 8K, and length-extrapolation PPL at
500M and 1.3B, and in LongBench at 1.3B.

\section*{Limitations}
\label{sec:limitations}

We evaluated CCQ only at 500M and 1.3B parameters with up to 40B
training tokens, only on two chunkwise linear-attention backbones
(GLA and Gated DeltaNet), and only on perplexity, S-NIAH,
length-extrapolation PPL, and LongBench; behaviour at 7B+ scale,
on Mamba/Mamba2-style selective state-space layers, with
RWKV-style channel mixing, or on agentic and multi-turn benchmarks
is not characterised.

CCQ assumes that retrieval-relevant keys lie outside the
high-variance directions of memory. If a useful target itself lies
along such a direction, as may occur for frequent patterns,
aggregation tasks, or templated targets, the correction can suppress
useful signal. Moreover, CCQ uses an undecayed covariance even though
GLA and Gated DeltaNet can forget or overwrite past information, so
$\Sigma_t$ need not match the backbone's effective memory. We leave
query-dependent, decay-aligned, and low-rank variants to future work;
Appendix~\ref{sec:app_choices} discusses these modelling choices.

The current implementation reduces training throughput by about
$16\%$, adds $8$--$10\%$ prefill latency and about $35\%$ decode
throughput loss, and increases recurrent state by $50\%$.

\section*{Acknowledgments}

This research is supported by the National Research Foundation, Singapore under its National Large Language Models Funding Initiative. (AISG Award No: AISG-NMLP-2024-005).

\bibliography{src/custom}

@inproceedings{transformer,
  title     = {Attention is All You Need},
  author    = {Vaswani, Ashish and Shazeer, Noam and Parmar, Niki and Uszkoreit, Jakob
               and Jones, Llion and Gomez, Aidan N. and Kaiser, {\L}ukasz and Polosukhin, Illia},
  booktitle = {Advances in Neural Information Processing Systems},
  volume    = {30},
  year      = {2017}
}

@inproceedings{linear_attn,
  title     = {Transformers are {RNN}s: Fast Autoregressive Transformers with Linear Attention},
  author    = {Katharopoulos, Angelos and Vyas, Apoorv and Pappas, Nikolaos and Fleuret, Fran{\c{c}}ois},
  booktitle = {Proceedings of the 37th International Conference on Machine Learning},
  series    = {Proceedings of Machine Learning Research},
  volume    = {119},
  pages     = {5156--5165},
  publisher = {PMLR},
  year      = {2020},
  url       = {https://proceedings.mlr.press/v119/katharopoulos20a.html}
}

@article{schmidhuber_fwp,
  title   = {Learning to Control Fast-Weight Memories: An Alternative to Dynamic Recurrent Networks},
  author  = {Schmidhuber, J{\"u}rgen},
  journal = {Neural Computation},
  volume  = {4},
  number  = {1},
  pages   = {131--139},
  year    = {1992},
  doi     = {10.1162/neco.1992.4.1.131}
}

@inproceedings{schlag_deltanet,
  title     = {Linear Transformers Are Secretly Fast Weight Programmers},
  author    = {Schlag, Imanol and Irie, Kazuki and Schmidhuber, J{\"u}rgen},
  booktitle = {Proceedings of the 38th International Conference on Machine Learning},
  series    = {Proceedings of Machine Learning Research},
  volume    = {139},
  pages     = {9355--9366},
  publisher = {PMLR},
  year      = {2021}
}

@inproceedings{deltanet,
  title     = {Parallelizing Linear Transformers with the Delta Rule over Sequence Length},
  author    = {Yang, Songlin and Wang, Bailin and Zhang, Yu and Shen, Yikang and Kim, Yoon},
  booktitle = {Advances in Neural Information Processing Systems},
  year      = {2024}
}

@inproceedings{gated_deltanet,
  title     = {Gated Delta Networks: Improving Mamba2 with Delta Rule},
  author    = {Yang, Songlin and Kautz, Jan and Hatamizadeh, Ali},
  booktitle = {The Thirteenth International Conference on Learning Representations},
  year      = {2025},
  url       = {https://openreview.net/forum?id=r8H7xhYPwz}
}

@inproceedings{mesanet,
  title     = {{MesaNet}: Sequence Modeling by Locally Optimal Test-Time Training},
  author    = {von Oswald, Johannes and Scherrer, Nino and Kobayashi, Seijin and Versari, Luca
               and Yang, Songlin and Schlegel, Maximilian and Maile, Kaitlin
               and Schimpf, Yanick and Sieberling, Oliver and Meulemans, Alexander
               and Saurous, Rif A. and Lajoie, Guillaume and Frenkel, Charlotte
               and Pascanu, Razvan and Ag{\"u}era y Arcas, Blaise and Sacramento, Jo{\~a}o},
  booktitle = {The Fourteenth International Conference on Learning Representations},
  year      = {2026},
  url       = {https://openreview.net/forum?id=xa3OnTb6c3}
}

@inproceedings{gla,
  title     = {Gated Linear Attention Transformers with Hardware-Efficient Training},
  author    = {Yang, Songlin and Wang, Bailin and Shen, Yikang and Panda, Rameswar and Kim, Yoon},
  booktitle = {Proceedings of the 41st International Conference on Machine Learning},
  series    = {Proceedings of Machine Learning Research},
  volume    = {235},
  publisher = {PMLR},
  year      = {2024}
}

@article{retnet,
  title   = {Retentive Network: A Successor to Transformer for Large Language Models},
  author  = {Sun, Yutao and Dong, Li and Huang, Shaohan and Ma, Shuming and Xia, Yuqing
             and Xue, Jilong and Wang, Jianyong and Wei, Furu},
  journal = {arXiv preprint arXiv:2307.08621},
  year    = {2023}
}

@inproceedings{rwkv,
  title     = {{RWKV}: Reinventing {RNN}s for the Transformer Era},
  author    = {Peng, Bo and Alcaide, Eric and Anthony, Quentin and Albalak, Alon
               and Arcadinho, Samuel and Biderman, Stella and Cao, Huanqi and Cheng, Xin
               and Chung, Michael and Grella, Matteo and GV, Kranthi Kiran and He, Xuzheng
               and Hou, Haowen and Lin, Jiaju and Kazienko, Przemyslaw and Kocon, Jan
               and Kong, Jiaming and Koptyra, Bartlomiej and Lau, Hayden and Mantri, Krishna Sri Ipsit
               and Mom, Ferdinand and Saito, Atsushi and Song, Guangyu and Tang, Xiangru
               and Wang, Bolun and Wind, Johan S. and Wozniak, Stanislaw and Zhang, Ruichong
               and Zhang, Zhenyuan and Zhao, Qihang and Zhou, Peng and Zhou, Qinghua
               and Zhu, Jian and Zhu, Rui-Jie},
  booktitle = {Findings of the Association for Computational Linguistics: EMNLP 2023},
  pages     = {14048--14077},
  publisher = {Association for Computational Linguistics},
  year      = {2023},
  doi       = {10.18653/v1/2023.findings-emnlp.936}
}

@inproceedings{mamba,
  title     = {Mamba: Linear-Time Sequence Modeling with Selective State Spaces},
  author    = {Gu, Albert and Dao, Tri},
  booktitle = {First Conference on Language Modeling},
  year      = {2024},
  url       = {https://openreview.net/forum?id=tEYskw1VY2}
}

@inproceedings{mamba2,
  title     = {Transformers are {SSM}s: Generalized Models and Efficient Algorithms Through Structured State Space Duality},
  author    = {Dao, Tri and Gu, Albert},
  booktitle = {Proceedings of the 41st International Conference on Machine Learning},
  series    = {Proceedings of Machine Learning Research},
  volume    = {235},
  pages     = {10041--10071},
  publisher = {PMLR},
  year      = {2024},
  url       = {https://proceedings.mlr.press/v235/dao24a.html}
}

@article{mongaras_softmax,
  title   = {On the Expressiveness of Softmax Attention: A Recurrent Neural Network Perspective},
  author  = {Mongaras, Gabriel and Larson, Eric C.},
  journal = {Transactions on Machine Learning Research},
  year    = {2025},
  url     = {https://openreview.net/forum?id=PHcITOi3vV}
}

@article{mamba2furious,
  title   = {{2Mamba2Furious}: Linear in Complexity, Competitive in Accuracy},
  author  = {Mongaras, Gabriel and Larson, Eric C.},
  journal = {arXiv preprint arXiv:2602.17363},
  year    = {2026},
  doi     = {10.48550/arXiv.2602.17363},
  url     = {https://arxiv.org/abs/2602.17363}
}

@inproceedings{performer,
  title     = {Rethinking Attention with Performers},
  author    = {Choromanski, Krzysztof and Likhosherstov, Valerii and Dohan, David and Song, Xingyou
               and Gane, Andreea and Sarl{\'o}s, Tam{\'a}s and Hawkins, Peter and Davis, Jared
               and Mohiuddin, Afroz and Kaiser, Lukasz and Belanger, David and Colwell, Lucy
               and Weller, Adrian},
  booktitle = {International Conference on Learning Representations},
  year      = {2021},
  url       = {https://openreview.net/forum?id=Ua6zuk0WRH}
}

@inproceedings{rfa,
  title     = {Random Feature Attention},
  author    = {Peng, Hao and Pappas, Nikolaos and Yogatama, Dani and Schwartz, Roy
               and Smith, Noah A. and Kong, Lingpeng},
  booktitle = {International Conference on Learning Representations},
  year      = {2021},
  url       = {https://openreview.net/forum?id=QtTKTdVrFBB}
}

@inproceedings{t2r,
  title     = {Finetuning Pretrained Transformers into {RNN}s},
  author    = {Kasai, Jungo and Peng, Hao and Zhang, Yizhe and Yogatama, Dani and Ilharco, Gabriel
               and Pappas, Nikolaos and Mao, Yi and Chen, Weizhu and Smith, Noah A.},
  booktitle = {Proceedings of the 2021 Conference on Empirical Methods in Natural Language Processing},
  pages     = {10630--10643},
  publisher = {Association for Computational Linguistics},
  year      = {2021},
  doi       = {10.18653/v1/2021.emnlp-main.830}
}

@inproceedings{hedgehog,
  title     = {The Hedgehog \& the Porcupine: Expressive Linear Attentions with Softmax Mimicry},
  author    = {Zhang, Michael and Bhatia, Kush and Kumbong, Hermann and R{\'e}, Christopher},
  booktitle = {The Twelfth International Conference on Learning Representations},
  year      = {2024},
  url       = {https://openreview.net/forum?id=4g02l2N2Nx}
}

@inproceedings{zoology,
  title     = {Zoology: Measuring and Improving Recall in Efficient Language Models},
  author    = {Arora, Simran and Eyuboglu, Sabri and Timalsina, Aman and Johnson, Isys
               and Poli, Michael and Zou, James and Rudra, Atri and R{\'e}, Christopher},
  booktitle = {International Conference on Learning Representations},
  year      = {2024},
  url       = {https://openreview.net/forum?id=LY3ukUANko}
}

@InProceedings{based,
  title = 	 {Simple linear attention language models balance the recall-throughput tradeoff},
  author =       {Arora, Simran and Eyuboglu, Sabri and Zhang, Michael and Timalsina, Aman and Alberti, Silas and Zou, James and Rudra, Atri and Re, Christopher},
  booktitle = 	 {Proceedings of the 41st International Conference on Machine Learning},
  pages = 	 {1763--1840},
  year = 	 {2024},
  editor = 	 {Salakhutdinov, Ruslan and Kolter, Zico and Heller, Katherine and Weller, Adrian and Oliver, Nuria and Scarlett, Jonathan and Berkenkamp, Felix},
  volume = 	 {235},
  series = 	 {Proceedings of Machine Learning Research},
  month = 	 {21--27 Jul},
  publisher =    {PMLR},
  url = 	 {https://proceedings.mlr.press/v235/arora24a.html}
}

@inproceedings{ruler,
  title     = {{RULER}: What's the Real Context Size of Your Long-Context Language Models?},
  author    = {Hsieh, Cheng-Ping and Sun, Simeng and Kriman, Samuel and Acharya, Shantanu
               and Rekesh, Dima and Jia, Fei and Ginsburg, Boris},
  booktitle = {First Conference on Language Modeling},
  year      = {2024},
  url       = {https://openreview.net/forum?id=kIoBbc76Sy}
}

@inproceedings{longbench,
    title = "{L}ong{B}ench: A Bilingual, Multitask Benchmark for Long Context Understanding",
    author = "Bai, Yushi  and
      Lv, Xin  and
      Zhang, Jiajie  and
      Lyu, Hongchang  and
      Tang, Jiankai  and
      Huang, Zhidian  and
      Du, Zhengxiao  and
      Liu, Xiao  and
      Zeng, Aohan  and
      Hou, Lei  and
      Dong, Yuxiao  and
      Tang, Jie  and
      Li, Juanzi",
    editor = "Ku, Lun-Wei  and
      Martins, Andre  and
      Srikumar, Vivek",
    booktitle = "Proceedings of the 62nd Annual Meeting of the Association for Computational Linguistics (Volume 1: Long Papers)",
    month = aug,
    year = "2024",
    address = "Bangkok, Thailand",
    publisher = "Association for Computational Linguistics",
    url = "https://aclanthology.org/2024.acl-long.172/",
    doi = "10.18653/v1/2024.acl-long.172",
    pages = "3119--3137"
}

@software{lm_eval_harness,
  author    = {Gao, Leo and Tow, Jonathan and Abbasi, Baber and Biderman, Stella
               and Black, Sid and DiPofi, Anthony and Foster, Charles and Golding, Laurence
               and Hsu, Jeffrey and Le Noac'h, Alain and Li, Haonan and McDonell, Kyle
               and Muennighoff, Niklas and Ociepa, Chris and Phang, Jason and Reynolds, Laria
               and Schoelkopf, Hailey and Skowron, Aviya and Sutawika, Lintang
               and Tang, Eric and Thite, Anish and Wang, Ben and Wang, Kevin and Zou, Andy},
  title     = {A framework for few-shot language model evaluation},
  year      = {2021},
  publisher = {Zenodo},
  doi       = {10.5281/zenodo.5371628},
  url       = {https://zenodo.org/records/5371628}
}

@inproceedings{fineweb,
  title     = {The {FineWeb} Datasets: Decanting the Web for the Finest Text Data at Scale},
  author    = {Penedo, Guilherme and Kydl{\'i}{\v{c}}ek, Hynek and Ben Allal, Loubna
               and Lozhkov, Anton and Mitchell, Margaret and Raffel, Colin
               and von Werra, Leandro and Wolf, Thomas},
  booktitle = {Advances in Neural Information Processing Systems},
  year      = {2024}
}

@inproceedings{ttt,
  title     = {Learning to ({L}earn at Test Time): {RNN}s with Expressive Hidden States},
  author    = {Sun, Yu and Li, Xinhao and Dalal, Karan and Xu, Jiarui and Vikram, Arjun
               and Zhang, Genghan and Dubois, Yann and Chen, Xinlei and Wang, Xiaolong
               and Koyejo, Sanmi and Hashimoto, Tatsunori and Guestrin, Carlos},
  booktitle = {Proceedings of the 42nd International Conference on Machine Learning},
  series    = {Proceedings of Machine Learning Research},
  volume    = {267},
  pages     = {57503--57522},
  publisher = {PMLR},
  year      = {2025},
  url       = {https://proceedings.mlr.press/v267/sun25h.html}
}

@inproceedings{titans,
 author = {Behrouz, Ali and Zhong, Peilin and Mirrokni, Vahab},
 booktitle = {Advances in Neural Information Processing Systems},
 doi = {10.52202/085713-3786},
 editor = {D. Belgrave and C. Zhang and H. Lin and R. Pascanu and P. Koniusz and M. Ghassemi and N. Chen},
 pages = {113506--113543},
 publisher = {Curran Associates, Inc.},
 title = {Titans: Learning to Memorize at Test Time},
 url = {https://proceedings.neurips.cc/paper_files/paper/2025/file/a4ca07aa108036f80cbb5b82285fd4b1-Paper-Conference.pdf},
 volume = {38, Main Conference},
 year = {2025}
}

@inproceedings{xlstm,
  title     = {{xLSTM}: Extended Long Short-Term Memory},
  author    = {Beck, Maximilian and P{\"o}ppel, Korbinian and Spanring, Markus and Auer, Andreas
               and Prudnikova, Oleksandra and Kopp, Michael and Klambauer, G{\"u}nter
               and Brandstetter, Johannes and Hochreiter, Sepp},
  booktitle = {Advances in Neural Information Processing Systems},
  year      = {2024}
}

@inproceedings{ramsauer_hopfield,
  title     = {Hopfield Networks is All You Need},
  author    = {Ramsauer, Hubert and Sch{\"a}fl, Bernhard and Lehner, Johannes
               and Seidl, Philipp and Widrich, Michael and Adler, Thomas and Gruber, Lukas
               and Holzleitner, Markus and Pavlovi{\'c}, Milena and Sandve, Geir Kjetil
               and Greiff, Victor and Kreil, David and Kopp, Michael and Klambauer, G{\"u}nter
               and Brandstetter, Johannes and Hochreiter, Sepp},
  booktitle = {International Conference on Learning Representations},
  year      = {2021},
  url       = {https://openreview.net/forum?id=tL89RnzIiCd}
}

@inproceedings{geva_ffn_keyvalue,
  title     = {Transformer Feed-Forward Layers Are Key-Value Memories},
  author    = {Geva, Mor and Schuster, Roei and Berant, Jonathan and Levy, Omer},
  booktitle = {Proceedings of the 2021 Conference on Empirical Methods in Natural Language Processing},
  pages     = {5484--5495},
  publisher = {Association for Computational Linguistics},
  year      = {2021},
  doi       = {10.18653/v1/2021.emnlp-main.446}
}

@article{expfam_var,
  title   = {Graphical Models, Exponential Families, and Variational Inference},
  author  = {Wainwright, Martin J. and Jordan, Michael I.},
  journal = {Foundations and Trends in Machine Learning},
  volume  = {1},
  number  = {1--2},
  pages   = {1--305},
  year    = {2008},
  doi     = {10.1561/2200000001}
}

\appendix

\section*{Appendix}
\label{sec:appendix}

\section{Derivations}
\label{sec:app_derivations}

This appendix collects the calculations referenced in
Section~\ref{sec:method}. The notation matches the main text: $q_t\in
\mathbb{R}^d$ is the query at step $t$, $\{k_m\}_{m\le t}$ are the
stored keys, and
$A_t(q) = \log\sum_{m\le t}\exp(k_m^\top q)$
is the softmax log-partition.

\subsection{Why \emph{curvature}?}
\label{sec:app_curvature}

The word \emph{curvature} in the title refers to the Hessian
$\nabla^2_q A_t(q)$ of the softmax log-partition. For a twice
differentiable function $f:\mathbb{R}^d\to\mathbb{R}$, the gradient
$\nabla f$ captures the local slope and the Hessian $\nabla^2 f$ the
local quadratic shape: directions where $\nabla^2 f$ is large are
directions in which $f$ rises steeply.

For $A_t$, this curvature has a concrete meaning: its Hessian is the
softmax-weighted covariance of the keys, so high-curvature directions
are directions of high key variation under the current attention
distribution. At $q=0$, this reduces to the unweighted running key
covariance. CCQ borrows only this local geometric signal and uses it
to contract the query at read time, without evaluating $\exp$.

\subsection{Taylor expansion of $A_t$}
\label{sec:app_taylor}

Expanding $A_t$ to second order around $q=0$ gives
\begin{equation}
\begin{aligned}
A_t(q) \;=\;\; & A_t(0) + \nabla A_t(0)^\top q \\
               & + \tfrac{1}{2}\, q^\top \nabla^2 A_t(0)\, q
                 + \mathcal{O}(\|q\|^3).
\end{aligned}
\end{equation}
A direct computation of $\nabla A_t(0)$ and $\nabla^2 A_t(0)$ proceeds
as follows. Define the softmax distribution
\begin{equation}
p_m(q) \;=\; \frac{\exp(k_m^\top q)}{\sum_{j\le t}\exp(k_j^\top q)},
\qquad m\le t,
\end{equation}
so that $A_t(q)$ is the log-normalizer of $p$. The standard exponential
family identities for the log-partition
function~\citep{expfam_var} give
\begin{align}
\nabla A_t(q)   &= \mathbb{E}_{p}[k]
               \;=\; \sum_{m\le t} p_m(q)\, k_m, \\
\nabla^2 A_t(q) &= \operatorname{Cov}_{p}[k] \notag \\
                &= \!\sum_{m\le t} p_m(q)\, k_m k_m^\top
                 - \mathbb{E}_p[k]\,\mathbb{E}_p[k]^{\!\top}.
\end{align}
At the isotropic-attention point $q=0$, every key receives equal
weight $p_m(0) = 1/t$, so
\begin{align}
\nabla A_t(0)   &\;=\; \mu_t \;=\; \tfrac{1}{t}\sum_{m\le t} k_m, \\
\nabla^2 A_t(0) &\;=\; \bar C_t - \mu_t\mu_t^\top
                 \;=\; \operatorname{Cov}_{\le t}[k],
\end{align}
which is Eq.~\eqref{eq:hess} in the main text. The two-term Hessian
is exactly the running sample covariance
$\Sigma_t \triangleq \bar C_t - \mu_t\mu_t^\top$ used by the cleaning
operator throughout the paper.

\subsection{Derivation of the cleaning operator}
\label{sec:app_newton}

The Hessian identity $\nabla^2_q A_t(0) = \Sigma_t$ supplies a
positive semidefinite operator whose eigendirections are precisely
the dominant covariance directions of memory. We use it to define a linear
\emph{cleaning operator} on the query space:
\begin{equation}
q_t^{\mathrm{clean}}
\;\triangleq\; (I - \lambda_t\, \Sigma_t)\, q_t
\;=\; q_t - \lambda_t\, \Sigma_t\, q_t,
\end{equation}
with a data-dependent gate $\lambda_t = \sigma(W_\lambda q_t +
b_\lambda)\in(0,1)$. Equivalently, defining
$R_t(q)=\tfrac{1}{2}q^\top\Sigma_tq$ gives
$q_t^{\mathrm{clean}}=q_t-\lambda_t\nabla_qR_t(q_t)$: CCQ is one
gradient step on the quadratic curvature term. Along
each eigendirection of $\Sigma_t$ with eigenvalue $\sigma_i$, the
operator multiplies the query component by $1-\lambda_t\sigma_i$, so
high-curvature directions of $A_t$ are contracted more strongly than
low-curvature ones, with $\lambda_t$ setting the overall strength.
The form $I-\lambda_t\Sigma_t$ realizes this anisotropic contraction
in the Hessian eigenbasis with a single scalar of freedom and is the
simplest first-order spectral contraction of this form.
Other spectral contractions, such as
$(I+\lambda_t\Sigma_t)^{-1}$ and $e^{-\lambda_t\Sigma_t}$, are also
possible and have $I-\lambda_t\Sigma_t$ as their first-order form, but
generally require a linear solve or additional matrix--vector products.
With unit-norm keys, $\|\Sigma_t\|_2 \le \|\bar C_t\|_2 \le 1$, so
the cleaning operator $I-\lambda_t\Sigma_t$ has spectrum in
$[1-\lambda_t,\, 1]\subset(0,1]$. No query direction is annihilated.
Since $\Sigma_t$ is symmetric positive semidefinite,
$I-\lambda_t\Sigma_t$ is itself symmetric and the cleaning is a soft
\emph{anisotropic contraction} in the eigenbasis of $\Sigma_t$; it
rescales eigen-directions but does not rotate them. This is why the
cleaning is stable at every position and at every sequence length,
and why CCQ behaves well even when $\lambda_t$ approaches 1.

This interpretation is local: the Hessian identity is exact at
$q=0$, while applying the marginal $\Sigma_t$ at trained queries is a
heuristic use of its eigendirections. Query/key normalization bounds
the correction but not the Taylor remainder. In particular, when the
true attention distribution concentrates on a few keys, its exact
query-weighted covariance can become low-rank and differ from the
global statistic used by CCQ; the learned gate can attenuate, but does
not eliminate, this mismatch.

\subsection{Score-space form of the correction}
\label{sec:app_score}

Substituting $q_t^{\mathrm{clean}} = q_t - \lambda_t \Sigma_t q_t$ into
the bilinear score $s_{tj} = k_j^\top q_t^{\mathrm{clean}}$ and
expanding $\Sigma_t = \bar C_t - \mu_t\mu_t^\top$,
\begin{align}
s_{tj}^{\mathrm{CCQ}}
&= k_j^\top q_t - \lambda_t\, k_j^\top \Sigma_t\, q_t \notag \\
&= k_j^\top q_t
 - \tfrac{\lambda_t}{t}\!\sum_{i\le t}\!(k_j^\top k_i)(k_i^\top q_t) \notag \\
&\phantom{= k_j^\top q_t}
 + \lambda_t\,(k_j^\top \mu_t)(\mu_t^\top q_t).
\end{align}
This is Eq.~\eqref{eq:score_form} in the main text. The first term is
the standard inner-product score. The second term penalizes a
candidate key $k_j$ when it aligns with a stored direction $k_i$
\emph{and} the query already activates that direction; the third term
adds back the rank-one contribution of the mean direction $\mu_t$, so
that only the spread around $\mu_t$ (not the mean itself) is
suppressed.

\section{Chunkwise computation of the running covariance}
\label{sec:app_chunkwise}

CCQ needs $\Sigma_t q_t = \bar C_t q_t - \mu_t (\mu_t^\top q_t)$ at
every position $t$. We compute the two terms by chunkwise scans that
share the same structure as the backbone's chunkwise update.

\paragraph{Second-moment term $\bar C_t q_t$.}
For training in the chunkwise mode of~\citet{deltanet,gla}, partition
the sequence into chunks $[1{:}C], [C{+}1{:}2C], \dots$ of fixed
length $C$. For a query at position $t$ in chunk $r$,
\begin{equation}
\bar C_t\, q_t
\;=\; \tfrac{1}{t}\,\bigl(S^{\mathrm{prev}}_{r} + I^{\mathrm{intra}}_{[r,t]}\bigr)\, q_t,
\end{equation}
where $S^{\mathrm{prev}}_{r} = \sum_{j\le rC} k_j k_j^\top$ is the
chunk-prefix sum and $I^{\mathrm{intra}}_{[r,t]} = \sum_{rC<j\le t} k_j k_j^\top$
is the intra-chunk contribution. $S^{\mathrm{prev}}_r$ has the form of
the linear-attention state for the kernel feature map $\phi(k) = k$
and the value stream taken to be the key stream; we maintain it as a
separate $d_k\!\times\!d_k$ running matrix alongside the backbone
state. The intra-chunk term is the dense
$Q_{[r]}\,(K_{[r]}^\top K_{[r]} \odot M)$-style block of \citet{gla}.
Length normalization by $1/t$ is applied after the product.

\paragraph{Mean term $\mu_t (\mu_t^\top q_t)$.}
The same chunk decomposition applies to the running mean:
$M^{\mathrm{prev}}_r = \sum_{j\le rC} k_j$ is carried as a
$d_k$-vector alongside $S^{\mathrm{prev}}_r$, and the intra-chunk
prefix sum
$I^{\mathrm{intra,1}}_{[r,t]} = \sum_{rC<j\le t} k_j$
is computed by a single cumulative sum within the chunk. Letting
$\tilde M_{[r,t]} \triangleq M^{\mathrm{prev}}_r + I^{\mathrm{intra,1}}_{[r,t]}$
denote the running key sum at position $t$, the mean correction
reduces to one inner product and one outer product per position:
\begin{equation}
\mu_t (\mu_t^\top q_t)
\;=\; \tfrac{1}{t^2}\, \tilde M_{[r,t]} \, \bigl(\tilde M_{[r,t]}^\top q_t\bigr).
\end{equation}
In practice both terms fuse with the backbone's existing chunkwise
kernel: one extra reduction per chunk materializes $S^{\mathrm{prev}}_r$
and $M^{\mathrm{prev}}_r$, and within the chunk we add one cumulative
sum, one inner product, and one outer product per query position.
Training throughput remains close to the underlying backbone.

\section{Algorithms}
\label{sec:app_algorithms}

Algorithm~\ref{alg:ccq_chunkwise} gives the chunkwise forward pass
used during training. The CCQ-specific lines are highlighted; the
rest of the loop is the backbone's chunkwise update and is left as an
opaque \textsc{BackboneUpdate} so the same procedure applies to GLA
and Gated DeltaNet.
Algorithm~\ref{alg:ccq_recurrent} gives the per-token recurrent
version used at inference time, with $\bar C_t$ maintained as a second
recurrent state alongside the backbone state $S_t$.

\begin{algorithm}[t]
\caption{CCQ chunkwise forward pass (training)}
\label{alg:ccq_chunkwise}
\begin{algorithmic}[1]
\Require sequence $\{x_t\}_{t=1}^{T}$, chunk size $C$, backbone state
         $S_0\!=\!0$, key second-moment state $S^{K}_0\!=\!0$, key mean state $M_0\!=\!0$, position $t\!\gets\!0$
\For{each chunk $r = 0, 1, \dots$ of length $C$}
  \State $Q_{[r]}, K_{[r]}, V_{[r]} \gets$ project $\{x_{rC+1},\dots,x_{(r+1)C}\}$
  \State $\bar Q_{[r]} \gets$ $\ell_2$-normalize $Q_{[r]}$
  \State $\bar K_{[r]} \gets$ $\ell_2$-normalize $K_{[r]}$
  \State \textcolor{blue}{\textbf{// CCQ read-side cleaning}}
  \State \textcolor{blue}{$P^{\mathrm{prev}} \gets S^{K}_{r}\, \bar Q_{[r]}^\top$
        \Comment{inter-chunk: prefix sum of $k k^\top$ times queries}}
  \State \textcolor{blue}{$P^{\mathrm{intra}} \gets (\bar Q_{[r]} \bar K_{[r]}^\top \odot M)\, \bar K_{[r]}$
        \Comment{intra-chunk dense block}}
  \State \textcolor{blue}{$\mathrm{CumK}_{[r]} \gets M_{r} + \textsc{cumsum}(\bar K_{[r]})$
        \Comment{running key sum at every position}}
  \State \textcolor{blue}{$D \gets \mathrm{diag}(1/(t+1),\, \dots,\, 1/(t+C))$
        \Comment{length normalization}}
  \State \textcolor{blue}{$\mathrm{MeanCorr}_{[r]} \gets D^2 \odot \mathrm{CumK}_{[r]}\, (\mathrm{CumK}_{[r]}^\top \bar Q_{[r]})$
        \Comment{$\mu_t (\mu_t^\top q_t)$, fused}}
  \State \textcolor{blue}{$\Lambda \gets \sigma(Q_{[r]}W_\lambda^\top + b_\lambda)$
        \Comment{per-token, per-head gates from pre-normalized queries}}
  \State \textcolor{blue}{$Q^{\mathrm{clean}}_{[r]} \gets \bar Q_{[r]} - \Lambda \odot \bigl(D\,(P^{\mathrm{prev}} + P^{\mathrm{intra}}) - \mathrm{MeanCorr}_{[r]}\bigr)$}
  \State $O_{[r]}, S_{r+1} \gets \textsc{BackboneUpdate}(Q^{\mathrm{clean}}_{[r]}, \bar K_{[r]}, V_{[r]}, S_{r})$
  \State \textcolor{blue}{$S^{K}_{r+1} \gets S^{K}_{r} + \bar K_{[r]}^\top \bar K_{[r]}$
        \Comment{update second-moment state}}
  \State \textcolor{blue}{$M_{r+1} \gets M_{r} + \sum_{j} \bar K_{[r],j}$
        \Comment{update mean state}}
  \State $t \gets t + C$
\EndFor
\State \Return $\{O_{[r]}\}$
\end{algorithmic}
\end{algorithm}

\begin{algorithm}[t]
\caption{CCQ recurrent inference (one token)}
\label{alg:ccq_recurrent}
\begin{algorithmic}[1]
\Require input $x_t$, backbone state $S_{t-1}$, second-moment state $\bar C_{t-1}$,
         mean state $\mu_{t-1}$, position $t$
\State $q_t, k_t, v_t \gets$ project $x_t$
\State $\bar q_t \gets q_t / \|q_t\|_2$
\State $\bar k_t \gets k_t / \|k_t\|_2$
\State \textcolor{blue}{$\bar C_t \gets \bar C_{t-1} + \tfrac{1}{t}(\bar k_t \bar k_t^\top - \bar C_{t-1})$
       \Comment{running-mean update of $kk^\top$}}
\State \textcolor{blue}{$\mu_t \gets \mu_{t-1} + \tfrac{1}{t}(\bar k_t - \mu_{t-1})$
       \Comment{running-mean update of $k$}}
\State \textcolor{blue}{$\Sigma_t \gets \bar C_t - \mu_t \mu_t^\top$
       \Comment{centred covariance / Hessian at $q\!=\!0$}}
\State \textcolor{blue}{$\lambda_t \gets \sigma(W_\lambda q_t + b_\lambda)$}
\State \textcolor{blue}{$q_t^{\mathrm{clean}} \gets \bar q_t - \lambda_t \Sigma_t \bar q_t$}
\State \small $o_t, S_t \gets \textsc{BackboneUpdate}\bigl($
        $q_t^{\mathrm{clean}}, \bar k_t, v_t, S_{t-1}\bigr)$
\State \Return $o_t$, new state $(S_t,\, \bar C_t,\, \mu_t)$
\end{algorithmic}
\end{algorithm}

In Alg.~\ref{alg:ccq_chunkwise}, the colour-highlighted lines are the
CCQ additions to the backbone forward pass; projections,
$\textsc{BackboneUpdate}$, and output computation follow the chosen
backbone. In Alg.~\ref{alg:ccq_recurrent}, the cache at inference time
is the triple $(S_t, \bar C_t, \mu_t)$: the backbone state, the running
key second moment, and the running key mean. They have sizes
$d_v d_k$, $d_k^2$, and $d_k$ respectively, so the total cache remains
$\mathcal{O}(1)$ in sequence length.

\paragraph{Wall-clock overhead.}
Table~\ref{tab:speed_overhead} reports the per-step wall-clock time
of a single attention/recurrent layer for each backbone and its CCQ
variant, measured on one H200 at the 1.3B layer shape
($H\!=\!2048$). We report forward $+$ backward at the 4K training
context (training-step cost) and forward-only at both the 4K
training context and 8K extrapolated context (inference-step cost).
Each entry is the mean $\pm$ standard deviation across 40
repetitions after 5 warm-up iterations, using
\texttt{torch.cuda.synchronize} around every step. The CCQ overhead
is $\sim$$1.2$\,ms on top of GLA and $\sim$$1.8$\,ms on top of Gated
DeltaNet for a 4K training step, an additive cost
dominated by the extra $d_k\!\times\!d_k$ statistic and the cleaning
projection. End-to-end measurements are reported below.

\begin{table}[t]
\centering
\small
\setlength{\tabcolsep}{2pt}
\begin{tabular}{lccc}
\toprule
Model & \multicolumn{1}{c}{fwd+bwd}
      & \multicolumn{2}{c}{forward only} \\
\cmidrule(lr){2-2}\cmidrule(lr){3-4}
 & 4K & 4K & 8K \\
\midrule
GLA                & $2.28\!\pm\!0.02$ & $0.75\!\pm\!0.02$ & $0.96\!\pm\!0.01$ \\
CCQ-GLA            & $3.50\!\pm\!0.03$ & $1.20\!\pm\!0.02$ & $1.37\!\pm\!0.01$ \\
Gated DeltaNet     & $6.26\!\pm\!0.02$ & $1.95\!\pm\!0.01$ & $2.98\!\pm\!0.03$ \\
CCQ-Gated DeltaNet & $8.06\!\pm\!0.03$ & $2.24\!\pm\!0.01$ & $3.67\!\pm\!0.03$ \\
\bottomrule
\end{tabular}
\caption{Per-layer wall-clock time (milliseconds) on one H200 at the
1.3B layer shape (batch 1, hidden 2048). \emph{fwd+bwd 4K} is the
training-step cost at the 4K context; \emph{forward only} is the
inference cost at the 4K training length and the 8K extrapolated
length. Each cell is mean $\pm$ standard deviation across 40
repetitions after 5 warm-up steps. CCQ adds $\sim$$1$--$2$\,ms
per 4K training step; forward overhead is reported at both lengths.}
\label{tab:speed_overhead}
\end{table}

\begin{figure*}[t]
\centering
\includegraphics[width=0.95\textwidth]{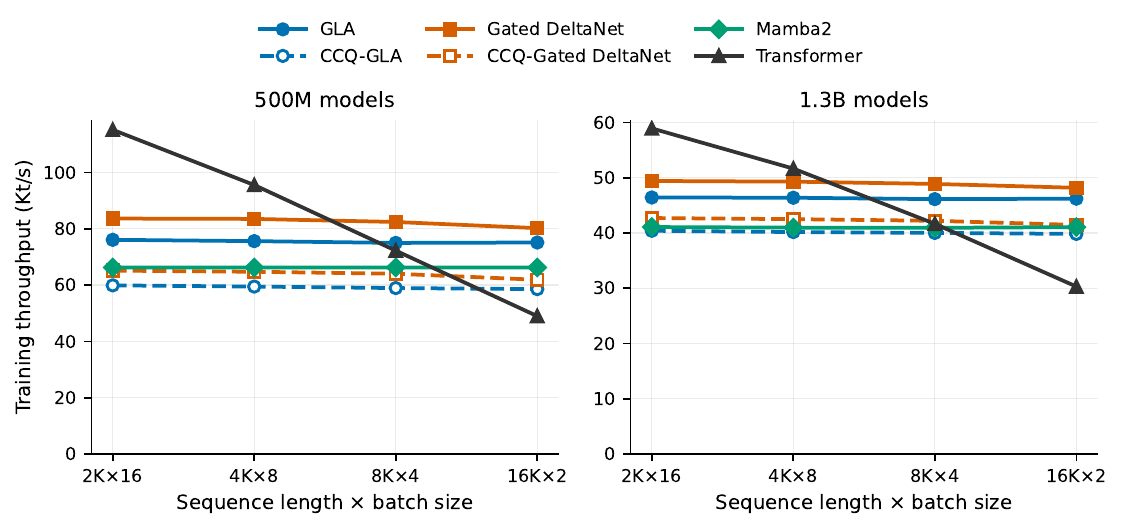}
\caption{Training throughput on one H200 at a constant 32K tokens per
step while varying sequence length and batch size. CCQ overhead is
approximately constant across sequence lengths; the Transformer
throughput falls as sequences grow.}
\label{fig:efficiency_overhead}
\end{figure*}

\paragraph{End-to-end efficiency.}
Figure~\ref{fig:efficiency_overhead} reports the controlled training
throughput sweep. Over the full 1.3B / 40B-token runs, median
throughput changes from 41.6k to 34.8k tokens/s/GPU for GLA and from
45.1k to 37.8k for Gated DeltaNet (about $16\%$), while peak training
memory rises from 87.5 to 99.9 GiB and from 81.5 to 93.6 GiB,
respectively. On one H200 with batch size 8, 4K prefill latency changes
from 198 to 217 ms for GLA and from 189 to 204 ms for Gated DeltaNet;
decode throughput changes from 454 to 298 and from 457 to 297 tokens/s.
The current decode path launches several additional small kernels per
layer; we do not assume an unimplemented fused CCQ decode kernel.

\section{Training Hyperparameters}
\label{sec:app_hparams}

Table~\ref{tab:hparams} reports the training hyperparameters used
across all runs. All variants share a 4K training context, SwiGLU
MLPs, RMSNorm, untied input/output embeddings, and the chunkwise
training mode; CCQ adds the read-side cleaning operator of
Sec.~\ref{sec:method_ccq} on top of its backbone. We use key/query
head dimension 256 and value head dimension 512 for GLA and Gated
DeltaNet at both scales. Queries and keys are $\ell_2$-normalized per
head in the CCQ variants. We inherit the remaining optimizer settings
of~\citet{gated_deltanet}. The CCQ-specific projection $W_\lambda$ is
a single linear layer producing one scalar gate per head. Its weights
are initialized to zero and its bias so that the initial gate is
approximately $0.01$. Per-model layer counts and token budgets are listed in
Table~\ref{tab:model_configs} of the main text.

\paragraph{Hardware.}
All models are trained on a single node with $8\times$ NVIDIA H200
GPUs using FSDP with mixed-precision bfloat16 and the chunkwise
training kernel of Sec.~\ref{sec:app_chunkwise}. The same node is
used for all evaluations (Sec.~\ref{sec:app_eval_details}): CSR/LM,
S-NIAH, length-extrap PPL, and LongBench. The CCQ-specific
$\Sigma_t q_t$ correction adds one $d_k\!\times\!d_k$ matrix and one
$d_k$-vector per head to the backbone state. This increases the
recurrent state by approximately $50\%$ (64 to 96 MiB per sequence in
the fp32 implementation), while remaining constant in sequence length.

\begin{table}[t]
\centering
\small
\setlength{\tabcolsep}{4pt}
\begin{tabular}{ll}
\toprule
Setting & Value \\
\midrule
Corpus & FineWeb-Edu (subset) \\
Tokenizer & 32K BPE \\
Context length & 4096 \\
Precision & bfloat16 \\
Optimizer & AdamW \\
$(\beta_1, \beta_2)$ & $(0.9,\, 0.95)$ \\
Weight decay & 0.1 \\
Peak learning rate & $10^{-3}$ \\
Warmup steps & 1024 \\
Schedule & cosine decay to 0 \\
\bottomrule
\end{tabular}
\caption{Training hyperparameters shared across all models.}
\label{tab:hparams}
\end{table}

\section{Covariance and Gate Ablations}
\label{sec:app_ablations}

We isolate covariance centring and the learned gate using the same
1.3B / 40B-token Gated DeltaNet recipe as the main experiments.

\begin{table*}[t]
\centering
\scriptsize
\setlength{\tabcolsep}{4pt}
\begin{tabular*}{\linewidth}{@{\extracolsep{\fill}}lrrrrrrrr}
\toprule
Variant & Wiki & LMB & Avg. & S1-4K & S1-8K & S2-4K & S2-8K & S3-4K \\
 & PPL$\downarrow$ & PPL$\downarrow$ & acc$\uparrow$ & acc & acc & acc & acc & acc \\
\midrule
Gated DeltaNet (no CCQ)                  & 17.98 & 13.11 & 55.64 & 100.0 & 99.8 & 83.4 & 56.6 & 6.0 \\
CCQ, centred $\Sigma_t$ (learned gate)  & 18.01 & 12.52 & 56.24 & 100.0 & 98.2 & 86.0 & 58.4 & 46.6 \\
CCQ, uncentred $\bar C_t$ (learned gate)& 17.30 & 12.55 & 56.20 & 100.0 & 97.6 & 84.8 & 56.2 & 55.6 \\
CCQ, centred $\Sigma_t$ (fixed $\lambda=0.1$) & 17.99 & 12.91 & 55.85 & 100.0 & 99.2 & 84.3 & 57.2 & 20.0 \\
\bottomrule
\end{tabular*}
\caption{Covariance and gate ablations at 1.3B / 40B tokens. S1--S3
denote the three S-NIAH tasks in Table~\ref{tab:s_niah}; Avg. is the
seven-task downstream average in Table~\ref{tab:results_lm}.}
\label{tab:cov_gate_ablation}
\end{table*}

Both covariance constructions improve over the backbone on several
metrics, showing that the gains do not depend on centring. The
uncentred variant is strongest on WikiText and S3-4K, while the centred
variant is stronger on number retrieval and the 8K cells; LAMBADA and
the downstream average remain similar. A fixed gate recovers part of
the gain but trails the learned gate most clearly on S3-4K.

\paragraph{What the learned gate captures.}
The current gate is a linear projection of the pre-split query and
does not observe the covariance directly. From an initial value of
approximately $0.01$, the sigmoid of its learned bias reaches
$0.35$--$0.45$ when averaged across layers. The realized per-token
gate averages approximately $0.29$ on ordinary text and rises to
$0.40$ at retrieval-query positions in single-needle prompts, averaged
over heads and layers. This input sensitivity is consistent with the
gate applying a stronger correction at retrieval queries, but does not
isolate why it helps. A more theory-aligned gate could use the exposure
$q_t^\top\Sigma_t q_t$ directly, while a retention-aligned variant
could couple its strength to the backbone's forget gate; we leave
matched comparisons of these designs to future work.

\paragraph{Inference-time intervention.}
For the trained CCQ-Gated DeltaNet checkpoint, disabling CCQ at
inference changes pass-key accuracy by at most 2.2 points from 4K to
100K and number/UUID retrieval by at most 1.6 points from 4K to 16K.
Thus, for this checkpoint, the gain over the separately trained
backbone persists without the correction at inference. This
intervention does not by itself identify the cause of the training
improvement.

\section{Evaluation Details}
\label{sec:app_eval_details}

This appendix expands the four evaluation families referenced in
Sec.~\ref{sec:exp_setup}.

\paragraph{(i) Common-sense reasoning and language modelling.}
We use \texttt{lm-evaluation-harness}~\citep{lm_eval_harness} for the
zero-shot tasks (no few-shot prompting, no in-context examples).
Reported tasks and metrics: PIQA (acc), HellaSwag (acc\_norm),
WinoGrande (acc), ARC-Easy (acc\_norm), ARC-Challenge (acc\_norm),
BoolQ (acc), SocialIQA (acc), LAMBADA OpenAI (perplexity and acc),
and WikiText word perplexity. The 7-task Avg column of
Table~\ref{tab:results_lm} is the unweighted mean of PIQA, HellaSwag,
WinoGrande, ARC-Easy, ARC-Challenge, SIQA, and BoolQ.

\paragraph{(ii) Synthetic in-context retrieval.}
Single-needle NIAH tasks of~\citet{ruler}, with
\texttt{max\_length}\,=\,$8{,}192$, batch size 8 at 1.3B and 32 at
500M. Table~\ref{tab:s_niah} resolves the three single-needle tasks
at 1K, 2K, 4K, and 8K (S-NIAH-1 pass-key, S-NIAH-2 number, S-NIAH-3
uuid; S-NIAH-3 stops at 4K because uuid retrieval collapses past
that for every model we tested).

\paragraph{(iii) Length-extrapolation perplexity.}
For each (dataset, length $L$) the dataset's text is tokenized into
a single stream, reshaped into non-overlapping windows of $L$
tokens, and PPL is reported as the exponential of the
token-weighted mean cross-entropy across windows. Nine evaluation
lengths from 4K to 20K in 2K steps on six long-context corpora
(GovReport, QMSum, NarrativeQA, Qasper, PG19, CodeParrot). All
models are trained at $L\!=\!4$K, so the 6K--20K points are pure
extrapolation. The figure in the main text
(Fig.~\ref{fig:length_extrap_500M}) shows the five recurrent
models whose perplexity is within a comparable range; the
Transformer and GLA-Hedgehog curves are off-range on every panel
and are not shown.

\paragraph{(iv) Long-context understanding (LongBench).}
The 14 English tasks of LongBench~\citep{longbench}: NarrativeQA,
Qasper, MultiField QA, HotpotQA, 2WikiMultiQA, MuSiQue (multi-doc QA);
GovReport, QMSum, MultiNews (summarisation); TREC (few-shot
classification); TriviaQA, SAMSum (single-doc QA, dialogue); LCC,
RepoBench-P (code completion). Each task uses its official metric
(F1 for QA, ROUGE-L for summarisation, classification accuracy for
TREC, code-edit similarity for the code tasks). The 14-task Avg is the unweighted
mean over the per-task scores. We only place the 1.3B block in the
main text because at the 500M scale the per-task standard error is comparable to the
between-model spread.

\section{Additional details on stability}
\label{sec:app_stability}

The cleaning operator $q \mapsto (I-\lambda_t\Sigma_t)q$ is only well
behaved when both $\Sigma_t$ and $\lambda_t$ are bounded. The two
design choices in Eq.~\eqref{eq:clean} guarantee this independently of
sequence length.

\textbf{Operator-norm bound on $\Sigma_t$.} For unit-norm keys, the
raw second moment $C_t = \sum_{j\le t} k_j k_j^\top$ has trace $t$,
so its operator norm grows linearly in $t$. The sample
second moment $\bar C_t = C_t/t$ has operator norm bounded by 1, since
\begin{align}
\|\bar C_t\|_2
&\;=\; \sup_{\|x\|=1}\, \tfrac{1}{t}\!\sum_{j\le t}\! (k_j^\top x)^2 \notag \\
&\;\le\; \sup_{\|x\|=1}\, \tfrac{1}{t}\!\sum_{j\le t}\! \|k_j\|^2 \|x\|^2
 \;=\; 1.
\end{align}
The centred covariance is dominated by the second moment:
$\Sigma_t = \bar C_t - \mu_t\mu_t^\top \preceq \bar C_t$, so
$\|\Sigma_t\|_2 \le \|\bar C_t\|_2 \le 1$.

\textbf{Spectrum of the cleaning operator.} With $\lambda_t\in(0,1)$
and $\|\Sigma_t\|_2 \le 1$, the operator $I-\lambda_t\Sigma_t$ has
spectrum in $[1-\lambda_t,\, 1] \subset (0, 1]$. Hence
$\|q_t^{\mathrm{clean}}\|_2 \le \|q_t\|_2$ and no direction is
annihilated. Bypassing either bound (using $C_t$ instead of
$\bar C_t$, or $\lambda_t$ without the sigmoid) breaks this guarantee
and we observed training instabilities on long sequences in early
ablations.

\section{Derivation of the retrieval-margin proposition}
\label{sec:app_proposition}

This appendix derives the retrieval-margin shift quoted in
Sec.~\ref{sec:method_ccq} as Eq.~\eqref{eq:margin_shift}, and
spells out the three interpretable regimes referenced in the body.

\paragraph{Setup.} Suppose the running key covariance is dominated by
a single high-variance direction $u\in\mathbb{R}^d$, so that
$\Sigma_t \approx \rho\, u u^\top$ for some $\rho\in(0,1]$. Let
$k_\star$ be a target key with $\alpha\!=\!\langle k_\star, u\rangle$,
let $k_d$ be a distractor key with $\beta\!=\!\langle k_d, u\rangle$,
and let $q$ be the current query.

\paragraph{Score difference.} The CCQ-corrected scores are
\begin{align}
s_\star^{\mathrm{CCQ}}
&= k_\star^\top q - \lambda\, k_\star^\top \Sigma_t\, q \notag \\
&\approx k_\star^\top q - \lambda\rho\, (k_\star^\top u)(u^\top q) \notag \\
&= k_\star^\top q - \lambda\rho\, \alpha\, (u^\top q),
\\
s_d^{\mathrm{CCQ}}
&= k_d^\top q - \lambda\, k_d^\top \Sigma_t\, q \notag \\
&\approx k_d^\top q - \lambda\rho\, \beta\, (u^\top q).
\end{align}
Subtracting gives the margin shift
\begin{equation}
s_\star^{\mathrm{CCQ}} - s_d^{\mathrm{CCQ}}
\;=\; (s_\star - s_d) + \lambda\rho\,(\beta - \alpha)\,(u^\top q).
\label{eq:margin_shift}
\end{equation}

\paragraph{Tight-cluster specialization.} If distractors lie tightly
along $u$, then $\beta\!\approx\!1$ and the margin shift simplifies to
$\lambda\rho\,(1-\alpha)\,(u^\top q)$.

\paragraph{Decomposing the query.} Writing $q = k_\star + \epsilon$
for a query that aims at the target plus some perturbation
$\epsilon$, we have $u^\top q = \alpha + u^\top\epsilon$, so the
margin shift becomes
\begin{equation}
\Delta \;=\; \lambda\rho\,(1-\alpha)\,(\alpha + u^\top\epsilon).
\end{equation}
The three regimes referenced in the body follow directly:

(i) If $\alpha<1$ and $\alpha + u^\top\epsilon > 0$, then $\Delta>0$
and CCQ widens the target margin. Concretely, this happens when the
target is not collinear with the high-variance direction and the query
has a positive overlap with it.

(ii) If $\alpha\!\to\!1$, the factor $(1-\alpha)$ vanishes and
$\Delta\!\to\!0$: the target lies along the high-variance direction,
so CCQ contracts target and distractors equally and the margin is
unchanged.

(iii) If $u^\top q\!\approx\!0$, then $\Delta\!\approx\!0$ regardless
of $\alpha$: the query has no projection along $u$, so the cleaning
operator has nothing to suppress in the first place.

\paragraph{Remarks.} The analysis is first-order in $\rho$ and ignores
sub-leading contributions from other eigenvectors of $\Sigma_t$ and
from distractors not collinear with $u$. Including those terms
preserves the qualitative picture: $\Delta$ has the same sign as
$(\beta-\alpha)(u^\top q)$ and vanishes at the same boundary cases.

\section{Empirical validation of the alignment assumption}
\label{sec:app_cov_validation}

The CCQ derivation assumes that retrieval-relevant keys live
\emph{off} the high-variance directions of memory, while distractor
keys cluster inside them. This appendix documents the diagnostic that
checks this assumption directly on pretrained models, as referenced
in Sec.~\ref{sec:method_properties} and Fig.~\ref{fig:cov_validation}.

\paragraph{Prompts.} To avoid drawing conclusions from a single
phrasing we use a collection of distinct needle scenarios that share
the same structural template but differ in the novel token, the
carrying sentence, and the query template (passwords, access codes,
magic words, courier phrases, parcel identifiers). Each prompt
concatenates $\approx\!120$ filler sentences drawn from a fixed pool
of generic English sentences, inserts the needle in the middle, and
ends with a question targeting the novel token. The distractor pool
is rotated between variants so each prompt sees a different filler
arrangement. After tokenization, each prompt is 1500--1700 tokens and
the needle span itself is 17--19 tokens.

A representative excerpt (with most distractors elided) is shown
below; \textbf{bold} marks the needle sentence and \emph{italic} marks
the trailing query:
\begin{quote}\small\sloppy
The cat sat on the mat and watched the rain through the window.
On Tuesday morning the conference room was unusually quiet.
She poured the coffee slowly, careful not to wake the dog.
$\langle$\dots\ $\sim\!60$ more filler sentences \dots$\rangle$\
\textbf{Please remember this very carefully: the secret password is
FLAMINGO47.}\
$\langle$\dots\ $\sim\!60$ more filler sentences \dots$\rangle$\
The librarian stamped the books and slid them across the counter.\
\emph{Q: What is the secret password? A: The secret password is}
\end{quote}

\paragraph{Unique-token filter.} The needle span necessarily includes
generic tokens (e.g.\ ``the'', ``is'', ``Please'') that also appear
all over the distractor context. Reporting the alignment over the full
needle span therefore mixes truly novel signal with generic
backbone tokens. We filter the needle set to token-ids that appear at
most twice in the full prompt, which keeps only the genuinely novel
sub-words (the FLAMINGO47, QUARTZ-2199 etc.\ pieces and a few rare
punctuation tokens). Fig.~\ref{fig:cov_validation} plots both the
unique-token mean (solid) and the raw all-needle mean (dashed) so the
effect of this filter is visible.

\paragraph{Targets.} Two models are probed.
\textbf{Qwen3-4B-Instruct-2507}
(softmax attention; 36 layers) is loaded in bfloat16.
\textbf{Gated DeltaNet 500M} (linear attention; 21 layers) is the
checkpoint we used as a baseline in Sec.~\ref{sec:exp_setup}. For each
model we sweep \emph{every} layer, since different layers play different
roles~\citep{geva_ffn_keyvalue} (early = lexicon / syntax, middle =
semantic content, late = task-specific routing) and a single-layer
report would not be representative of the whole network.

\paragraph{Capture procedure.} A forward hook is attached to the
selected layer's $W_K$ projection. After one forward pass over the
needle-in-a-haystack prompt, the captured tensor is reshaped to
$(T, H, d_k)$ where $T$ is the token length and $H$ is the number of
key heads (with grouped-query attention, $H$ is the number of \emph{kv}
heads). Each per-head key vector is then $\ell_2$-normalized to match
CCQ's standing assumption on the key magnitudes.

\paragraph{Alignment measure.} For each head we compute the centred
covariance
$\Sigma^{(h)} = \tfrac{1}{T-1}\sum_t (k^{(h)}_t - \bar k^{(h)})(k^{(h)}_t - \bar k^{(h)})^\top$
and its top-16 eigenvectors $U^{(h)} \in \mathbb{R}^{d_k\times 16}$
(by eigh on a symmetric $d_k\!\times\!d_k$ matrix, $d_k\in\{128,256\}$
for our targets). For every token $t$, the head-averaged
\emph{top-16 alignment} is
\begin{equation}
a^{(h)}_t \;=\; \frac{\|U^{(h)\top}(k^{(h)}_t - \bar k^{(h)})\|^2}
                        {\|k^{(h)}_t - \bar k^{(h)}\|^2}\,,
\end{equation}
the fraction of the centred key's energy that sits in the high-variance
subspace. We then average $a^{(h)}_t$ separately over (i) the tokens
that fall inside the needle sentence and (ii) the remaining context
tokens (excluding the trailing query tokens), obtaining the needle
mean $\mu_{\text{needle}}$ and distractor mean $\mu_{\text{distractor}}$
for that head and layer. The \emph{alignment gap} reported throughout
this section is their signed difference,
\begin{equation}
\Delta_\mu \;\triangleq\; \mu_{\text{needle}} - \mu_{\text{distractor}},
\end{equation}
so $\Delta_\mu < 0$ means needle tokens project \emph{less} onto the
top-16 high-variance subspace than distractor tokens do — the
geometric configuration CCQ assumes. $|\Delta_\mu|$ measures how
strongly the two groups are separated, and the sign indicates which
group lives off the high-variance axes. The per-layer values plotted
in Fig.~\ref{fig:cov_validation} are $\Delta_\mu$ averaged across
heads.

\paragraph{Results.} Fig.~\ref{fig:cov_validation} shows
$\Delta_\mu$ for every layer of each model, averaged over the
prompt collection. The unique-token $\Delta_\mu$ is \textbf{negative
at every single layer of both models} (36/36 for Qwen3, 21/21 for Gated
DeltaNet) with a per-layer prompt-wise standard deviation around
$0.02$. The overall depth-averaged separation is
$\bar\Delta_\mu = -0.111$ for Qwen3-4B-Instruct and
$\bar\Delta_\mu = -0.209$ for Gated DeltaNet, roughly $2\times$ larger
in the linear-attention model. Both curves share the same depth
shape: strongest in the earliest layers, weakest near the middle, and
partially recovering in the later layers, without ever crossing
zero.

The depth shape needs a careful reading because the three regions
have different mechanisms behind them, consistent with known
transformer-layer specialization~\citep{geva_ffn_keyvalue}. At
\emph{early layers} keys are still close to token embeddings, so a
rare token like \texttt{FLAMINGO47} occupies its own near-orthogonal
direction simply by virtue of being rare — and any rare token would
do the same. The strong negative $\Delta_\mu$ at depth 0--3 is
therefore partly a \emph{lexical-rarity} signal and is \emph{not} by
itself strong evidence for the CCQ premise. The middle layers are the
more demanding test: there each key has been integrated with its
sentence context, so token identity has been smeared into shared
semantic directions and the early-layer rarity artefact is gone.
Despite that smearing, $\Delta_\mu$ is still negative in the middle
($\approx\!-0.05$ for Qwen3, $\approx\!-0.20$ for Gated DeltaNet); the
needle key continues to sit off the high-variance subspace of memory
even when it shares semantic content with the surrounding sentences.
This is the regime CCQ is designed to act on. The partial recovery of
$|\Delta_\mu|$ in the late layers is consistent with task-specific
features re-separating the novel token at the answer position.

Fig.~\ref{fig:cov_validation_distribution} additionally plots the
per-token distribution of alignment values, pooled over every layer
and prompt. The distractor and unique-needle histograms are clearly
separated in both models, providing a direct geometric visualization
of the statistic that Fig.~\ref{fig:cov_validation} resolves by depth.

\begin{figure*}[t]
\centering
\includegraphics[width=0.9\textwidth]{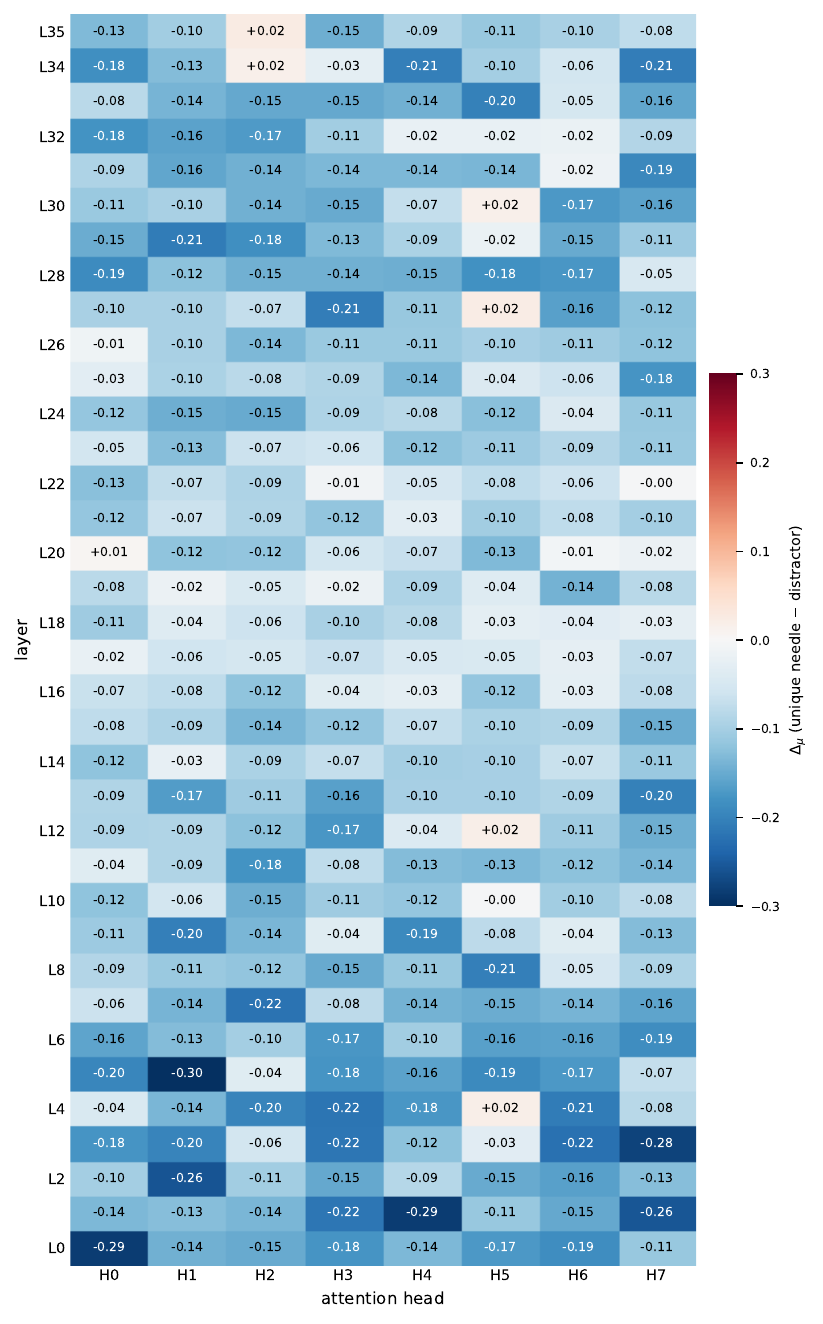}
\caption{Per-(layer, head) $\Delta_\mu$ heatmap for
Qwen3-4B-Instruct (36 layers $\times$ 8 key heads). Each cell is
$\Delta_\mu = \mu_{\text{unique-needle}} - \mu_{\text{distractor}}$
on top-16 alignment, pooled across all prompts. Diverging colormap
centred at zero; blue = negative (matches the CCQ premise), red =
positive. Annotated value is $\Delta_\mu$ for that cell.}
\label{fig:cov_validation_per_layer_head_qwen3}
\end{figure*}

\begin{figure*}[t]
\centering
\includegraphics[width=0.9\textwidth]{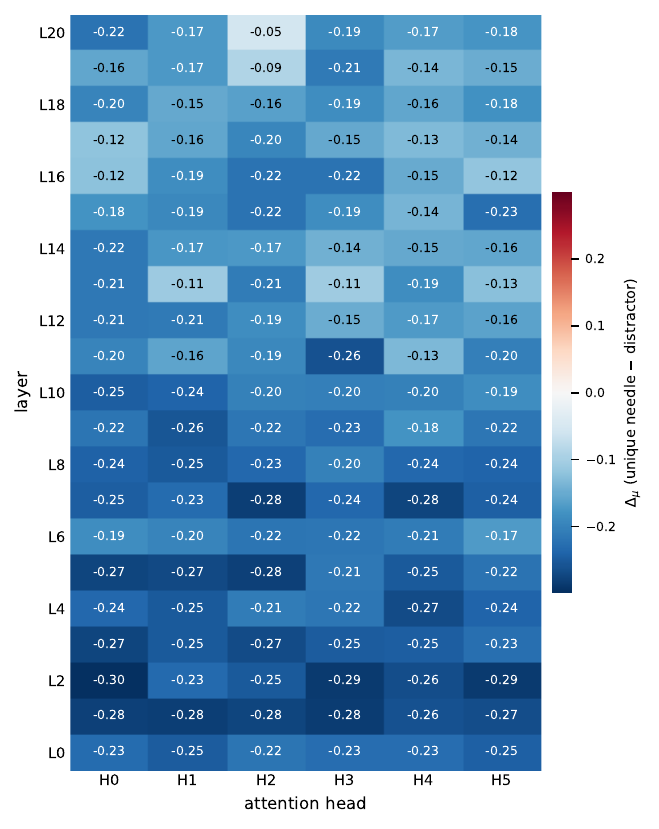}
\caption{Per-(layer, head) $\Delta_\mu$ heatmap for Gated DeltaNet
500M (21 layers $\times$ 6 heads). Same colormap and pooling
convention as Fig.~\ref{fig:cov_validation_per_layer_head_qwen3}.
The separation is visibly larger and more uniform than in the
softmax model: every (layer, head) cell is strongly negative.}
\label{fig:cov_validation_per_layer_head_gd}
\end{figure*}

\paragraph{Per-(layer, head) breakdown.}
Figures~\ref{fig:cov_validation_per_layer_head_qwen3}
and~\ref{fig:cov_validation_per_layer_head_gd} resolve the same
statistic at the finest granularity available: one cell per
\emph{(layer, head)} pair, with no averaging across layers. Each
heatmap has rows indexing layers (top = deepest, bottom =
shallowest) and columns indexing attention heads. The cell colour and
annotated value report $\Delta_\mu$ (unique-needle mean minus
distractor mean) for that specific (layer, head) pair, pooled across
all prompts. Blue cells (the dominant colour in both models)
indicate the configuration CCQ assumes; red cells indicate the
opposite. The pooled-across-cells distribution shape itself is shown
in Fig.~\ref{fig:cov_validation_distribution}, so the heatmap focuses
exclusively on the separation statistic.

The geometric premise --- needle keys sit \emph{off} the
high-variance subspace of memory --- holds at the cell level, not
just on average. Almost every $(\ell, h)$ cell in both models
shows $\Delta_\mu < 0$; the few cells with $\Delta_\mu \gtrsim 0$
are isolated and concentrated in the small handful of mid-network
layers where the depth-averaged separation in
Fig.~\ref{fig:cov_validation} is at its weakest. The linear-attention
model continues to show a uniformly larger separation than the
softmax model, head by head and layer by layer. This per-cell
consistency is the strongest evidence we can offer for the claim
short of training a new model.

\paragraph{Interpretation.} The signal is in the right direction at all
probed layers of both architectures. The fact that linear-attention
keys show a stronger separation matches the intuition that linear
attention has no softmax denominator to lean on and must learn this
distinction in the key space itself. The diagnostic does not by
itself prove that CCQ improves retrieval (only the
language-modelling, S-NIAH, length-extrapolation, and LongBench
tables in Sec.~\ref{sec:exp_lm}, \ref{sec:exp_s_niah},
\ref{sec:exp_length_extrap}, and~\ref{sec:exp_longbench} can do
that), but it shows that the underlying geometric premise of the
derivation in Sec.~\ref{sec:method_properties} is realized in
trained models rather than only in the toy retrieval analysis.

\section{Modelling choices and limitations}
\label{sec:app_choices}

We collect here the modelling choices that simplify CCQ's local
quadratic model, and the limitations they imply.

\paragraph{Why we use the unweighted covariance.}
The exact softmax Hessian away from $q\!=\!0$ is the
\emph{$p(q)$-weighted} covariance $\operatorname{Cov}_{p(q)}[k]$, which
re-weights keys by their current attention mass. CCQ approximates this
$q$-dependent operator by the unweighted running covariance $\Sigma_t$,
gaining a pair of shared running statistics per head and avoiding any
$\exp(\cdot)$ evaluation. The cost is that variance is measured over
the marginal distribution of past keys, not over the conditional
distribution induced by the current query; we view this as a coarse
but cheap surrogate, and the empirical results suggest it is faithful
enough for retrieval-heavy tasks. A $q$-dependent extension (e.g.\ a
low-rank local reweighting of $\Sigma_t$ toward
$\operatorname{Cov}_{p(q)}[k]$) is a natural follow-up.

\paragraph{Interaction with decayed backbones.}
When the backbone forgets, the effective memory read from $S_t$ is not
exactly $\sum v_j k_j^\top$ but a decayed sum
$\sum \gamma_{j,t} v_j k_j^\top$ for some per-key factor
$\gamma_{j,t}\in(0,1]$. CCQ uses an \emph{undecayed} covariance
$\Sigma_t$, which can over-penalize directions that the backbone has
already forgotten. We did not observe a systematic degradation at
our scales: CCQ-Gated DeltaNet improves WikiText perplexity at 500M
and average downstream accuracy at both scales, while its 1.3B
WikiText perplexity is essentially unchanged (18.01 vs.\ 17.98). A
decay-aligned covariance $\Sigma_t^{\gamma}$ computed from
$\gamma$-weighted running statistics
is a principled variant and a useful direction for future work,
particularly for backbones with aggressive forgetting.

\end{document}